\documentclass[twoside]{article}

\usepackage[accepted]{aistats2022}
\usepackage[round]{natbib}

\usepackage{booktabs}       
\usepackage{amsfonts}       
\usepackage{nicefrac}       
\usepackage{microtype}      
\usepackage{xcolor}         

\usepackage{graphicx}
\usepackage{float}
\usepackage[caption = false]{subfig}
\usepackage{amsmath}
\usepackage{comment}

\usepackage{amssymb}
\usepackage{bbm}
\usepackage{multirow}
\usepackage[ruled,vlined]{algorithm2e}
\usepackage{tablefootnote}
\newcommand{\clusters}{C}

\usepackage[font=small,labelfont=bf]{caption}

\begin{document}

%

%

\twocolumn[

\aistatstitle{TD-GEN: Graph Generation Using Tree Decomposition}

\runningtitle{TD-GEN}

\runningauthor{Hamed Shirzad, Hossein Hajimirsadeghi, Amir H. Abdi, Greg Mori}

\aistatsauthor{ Hamed Shirzad \And Hossein Hajimirsadeghi }

\aistatsaddress{ Borealis AI \& Simon Fraser University  \And  Borealis AI }

\aistatsauthor{Amir H. Abdi \And Greg Mori}
\aistatsaddress{ Borealis AI \And Borealis AI \& Simon Fraser University}

]

\begin{abstract}
  We propose TD-GEN, a graph generation framework based on tree decomposition, and introduce a reduced upper bound on the maximum number of decisions needed for graph generation. The framework includes a permutation invariant tree generation model which forms the backbone of graph generation. Tree nodes are supernodes, each representing a cluster of nodes in the graph. Graph nodes and edges are incrementally generated inside the clusters by traversing the tree supernodes, respecting the structure of the tree decomposition, and following node sharing decisions between the clusters. Further, we discuss the shortcomings of the standard evaluation criteria based on statistical properties of the generated graphs.  We propose to compare the generalizability of models based on expected likelihood. Empirical results on a variety of standard graph generation datasets demonstrate the superior performance of our method.
\end{abstract}

\section{INTRODUCTION}
\label{sec:intro}

Graph generation using deep generative models is an active area of research, and recent probabilistic approaches based on variational autoendoers~\citep{simonovsky2018graphvae}, autoregressive models~\citep{you2018graph, liao2019efficient, dai2020scalable}, and normalizing flows~\citep{liu2019graph, shi2020graphaf} have provided high-capacity models to learn graph distributions. In this work, we propose a general framework for generic graph generation by learning a distribution over graphs. However, our main goal is to have better generalizability on unseen data with moderately efficient graph generation steps.

Compared to unstructured data generation (e.g., images), graph generation poses new challenges 
such as isomorphism and non-unique representations. This makes node matching in reconstruction-based models such as GraphVAE~\citep{simonovsky2018graphvae} complex.  Recent autoregressive models~\citep{you2018graph, shi2020graphaf, su2019graph, hajiramezanali2019variational, chen2021order} try to tackle this by sequential graph generation using node ordering. However, because of the factorial number of possible node permutations, the hypothesis space of graph representations grows super-exponentially with the number of nodes.  Without significantly reducing the hypothesis space, the models either underfit the true distribution or overfit to the permutations seen during training. This issue has been overlooked in prior work, and we show that the statistics-based performance measures commonly used for graph generation are misleading in this respect.

Tree decomposition transforms a graph into a tree, where each node of the tree is a subgraph of the main graph. The width of a tree decomposition is the number of nodes in the largest such subgraph.  There can be multiple tree decompositions for a given graph; the treewidth of a graph is the minimum width among all possible tree decompositions. 
Tree decomposition can facilitate breaking down graphs into smaller components and helps to tackle the aforementioned problems with reduced complexity. This method has been previously used in graphical models~\citep{murphy2012machine} and molecule generation~\citep{jin2018junction}.  The treewidth of many real-world graphs is much smaller than the number of nodes in the graph (e.g., ~\citep{maniu2019experimental}).

Our proposed graph generation is based on tree decomposition. The tree provides an abstract representation of a graph. Each node of the tree, i.e., supernode, contains a subgraph of the main graph. Thus, graph generation is reduced to the generation of the tree structure followed by subgraph generation inside each supernode. This tree decomposition is beneficial to address various aspects of graph generation. It can reduce the hypothesis space of graph representations and consequently mitigate the challenges of isomorphisms and non-unique representations. It also helps simplify learning local relations in the graph by focusing on learning connections inside each subgraph. On the other hand, global relations in the graph are captured via the tree representation, which is simpler than the original graph. For this, we propose a novel permutation invariant tree representation which leads to a permutation invariant tree generator. 
In aggregate, the proposed framework can reduce the number of decision steps for graph generation from $O(n^2)$ to $O(nk)$ where $k$ is the width of the tree decomposition.

The contributions of this paper are as follows: 
(1) A novel formulation for graph generation based on tree decomposition, which can alleviate existing challenges such as the 
multitude of generation steps, non-unique graph representations, and intractable hypothesis spaces.  
(2) A permutation invariant model for tree generation based on a new tree representation, which reduces the hypothesis space by eliminating permutations and outperforms the state-of-the-art graph generation methods.
(3)~Proofs of an upper bound on the number of decisions required for graph generation.
(4) Demonstrating the inefficacy of the statistics-based metrics for evaluating graph generation models and suggesting reusing expected likelihood over different orderings as a more indicative metric for generalizability.

\section{RELATED WORK}
\label{relWork}
The graph counterparts of modern generative models have received significant attention in the past few years.
GraphVAE~\citep{simonovsky2018graphvae} is an extension of the variational autoencoder~\citep{kingma2013auto}, which generates whole adjacency matrix ($O(n^2)$ parameters) for small graphs. Algorithm requires $O(n^4)$ approximation algorithm for matching input and output graphs.
NetGAN extends generative adversarial networks (GAN)~\citep{goodfellow2014generative} with a Wasserstein objective to learn the distribution of random walks over graphs~\citep{bojchevski2018netgan}. 
GraphRNN is an autoregressive model which uses recurrent neural networks to generate graphs from a sequential ordering of the nodes~\citep{you2018graph}.
GraphVRNN and VGRNN are variational extensions of GraphRNN, which leverage latent factors to capture more complex dynamics, variability, and uncertainty~\citep{su2019graph,hajiramezanali2019variational}. 
GraphAF is an autoregressive graph generation model which combines the advantages of sequential models and normalizing flows for high capacity density estimation~\citep{shi2020graphaf}.
GNF and EDP-GNN both use Graph Neural Networks (GNN) for permutation invariant generation of graphs~\citep{liu2019graph, niu2020permutation}. GNF uses reversible GNNs based on normalizing flows, and EDP-GNN uses score matching for the graph generation.

Various research has been also conducted on the molecule graph generation task~\citep{jin2018junction,you2018graphConv, shi2020graphaf,jin2020hierarchical,zang2020moflow,polishchuk2020crem}. Considering well-known motifs and structures in the molecules and the domain knowledge (\emph{e.g.}, the degree of nodes and molecule validity) give a huge advantage to this domain-specific task. Also, in learning a distribution over a single temporal large graph~\citep{pennycuff2017temporal} has previously used tree decomposition, this work uses graph grammars for learning the distribution. However, the focus of our work is generic graph generation, where no prior information is available, and the model learns the whole process from scratch to learn a distribution on a dataset of several graphs.

In this work, we propose a novel framework for general graph generation using tree decomposition. A seminal work leveraging tree structures for graph generation was done in JT-VAE~\citep{jin2018junction}. But, JT-VAE was proposed for molecule generation and simply uses a vocabulary of predetermined structures (\emph{e.g.}  rings, bonds, and atoms) to convert graph-structured molecules to tree-structured scaffolds. Our work is different from JT-VAE in multiple aspects: first, our proposed model is generic and has no requirement of predetermined structures. This positively impacts the flexibility of the framework. Second, we mathematically formulate how graph generation can be decomposed into tree generation followed by subgraph generation and provide proofs on how this will reduce the number of decision steps for graph generation. Third, we propose a novel permutation invariant tree generator, which provides representational efficiency. 

\section{BACKGROUND}
\label{background}

\textbf{Tree Decomposition:} For graph $G = (\mathcal{V}, \mathcal{E})$, let $\mathcal{T} = (T,\clusters)$, where $T$ is a tree with $r$ nodes, and $\clusters = \{C_1,C_2,\ldots,C_r \}$ are non-empty induced subgraphs of $G$, mapped to the nodes of the $T$ (we name these nodes as supernodes). We define $\mathcal{X} = \{ X_1, X_2, \ldots, X_r\}$, where $X_i$ denotes the set of nodes in $C_i$, which we refer to as bags. 
Accordingly, $\mathcal{T}$ is a tree decomposition of $G$ if it meets the following conditions:
\begin{enumerate}
\setlength\itemsep{0cm}
    \item $X_1 \cup X_2 \cup \cdots \cup X_r = \mathcal{V}$,
    \item $\forall e=(u,v) \in \mathcal{E} 	\quad \exists X_i:u,v \in X_i$,
    \item For a node $v$, where 
    $v \in X_i$ and $v \in X_j$, 
    for each supernode $k$ in the path between supernode $i$ and supernode $j$ in $T$, $v \in X_k$. 
    In other words, the induced subgraph of $T$ containing the node $v$ in their corresponding bags is a connected graph.
\end{enumerate}
Knowing $T$ and $\clusters$, $G$ is uniquely determined; however, for a given graph the tree decomposition is not unique.

\textbf{Width and Treewidth: } The width $w_{\mathcal{T}}$ of a tree decomposition $\mathcal{T}$ is the size of its largest bag, which can be formulated as $w_{\mathcal{T}} = max_i |X_i|$. The minimum width among all of possible tree decompositions of graph $G$, i.e., $\mathcal{D}(G)$, is called the treewidth of $G$.
Treewidth of $G$ is formulated as
$tw(G) := 
min_{\mathcal{T} \in \mathcal{D}(G)} 
w_{\mathcal{T}}$. Finding the treewidth is an NP-hard problem in general. However, there are approximate solutions for it~\citep{bodlaender2010treewidth}. 


\section{PROPOSED METHOD}
\label{sec:method}
\begin{figure}[t!]
  \centering
  \includegraphics[width=80mm]{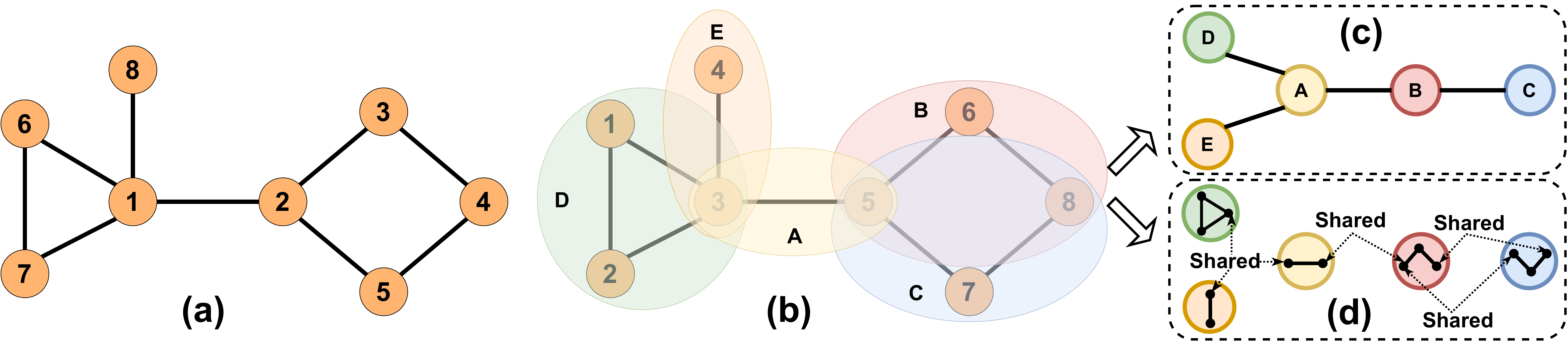}
  \caption{Overview of the proposed framework. (a) Input graph $G$. (b) Tree decomposition of $G$, where each ellipse shows a cluster of nodes corresponding to a tree node (supernode). As a result, graph generation is decomposed into tree generation and subgraph generation for each cluster.
  (c) The tree structure $T$, for which we propose a permutation-invariant generative model. 
  (d) Subgraphs from the tree decomposition, which are generated individually within two steps of node sharing with the parent supernode and new node generation.
  }
  \label{fig:method-overview}
  \vspace{-0.1in}
\end{figure}

A high-level overview of our proposed method is shown in Figure~\ref{fig:method-overview}. Given an undirected graph $G$, a minimal tree decomposition algorithm is used to decompose the graph into a tree $T$ and a set of subgraphs $\clusters = \{C_1,\ldots, C_r\}$ corresponding to each node of the tree. Using tree decomposition, graph generation can be reduced to learning two models, one for tree structure generation and one for generating the subgraphs. This reduction has multiple benefits. First, it can take advantage of efficient message passing in trees for encoding and representation. Further, it can alleviate some of the challenges in graph generation. While graph generation generally deals with non-unique representations with factorial growth of node permutations, we propose a permutation invariant tree generator based on an isomorphic representation of the tree. On the other hand, we show how the generation of each new subgraph is simplified into two steps: node sharing with the parent subgraph and generating the remaining nodes of the new subgraph. Thus, the decision space is confined to the nodes of the new subgraph and the parent subgraph (\emph{i.e.} $O(w_{\mathcal{T}})$), which can be significantly smaller than the total number of graph nodes in practice (see Appendix \ref{appendix:stats}).

\subsection{Model Formulation}
\label{sec:model-formulation}

Graph generation can be formulated as learning a distribution $p(G)$ over graphs. Prior autoregressive models such as GraphRNN~\citep{you2018graph} and its successive extensions ~\citep{liao2019efficient, su2019graph, hajiramezanali2019variational} aim to model $p(G)$ as the distribution of the adjacency matrix $\boldsymbol{A}$ under different node orderings $\pi$:
\begin{equation}
p(G) = \sum_{\pi \in \Pi}{p(G,\pi)} = \sum_{\pi \in \Pi_{\mathrm{Adj}}}{p(\boldsymbol{A}^{\pi})},
\end{equation}
where $\Pi_{\mathrm{Adj}}$ is the set of permutations with unique adjacency matrices, which is a subset of all possible 
$n!$ permutations $\Pi$.
As a result, learning the graph distribution is reduced to learning a distribution over distinctive adjacency matrices of the graph, where each adjacency matrix $\boldsymbol{A}^{\pi}$ is further represented as a sequence of its rows $S_i^{\pi}$ according to $\pi$ and induced by the mapping $f_S$~\citep{you2018graph}:
$$(S_1^{\pi}, \cdots, S_n^{\pi}) = f_{S}(G, \pi).$$
Here, we propose a novel graph generation approach by introducing the mapping $f_{T}$ from graphs to tree decompositions: 
$(T^{\pi}, \clusters^{\pi}) = f_{T}(G, \pi)$.
Thus, the graph distribution can be formulated as
\begin{equation}
\label{eq:td-graph-prob}
p(G) = \sum_{\pi \in \Pi_{\mathrm{TD}}}{p(f_{T}(G, \pi))} = \sum_{\pi \in \Pi_{\mathrm{TD}}}{p(T^{\pi}, \clusters^{\pi})},
\end{equation}
where $\Pi_{\mathrm{TD}}$ is the set of node permutations with distinctive tree decompositions of $G$. We experimentally show that size of $\Pi_{\mathrm{TD}}$ can be significantly smaller than $n!$. In fact, without having a proper space reduction in mapping, learning the true distribution of graphs is intractable. In Section~\ref{sec:learning-complexity}, we expand on this issue, which has been overlooked in the prior work.

Given the formulation for $p(G)$ in Eq.~\ref{eq:td-graph-prob}, the goal is to model $p(T^{\pi}, \clusters^{\pi})$, 
\emph{i.e.}, the joint distribution of the tree $T^{\pi}$ and its subgraphs $\clusters^{\pi}$ under the node ordering $\pi$, which can be written as $p(T^{\pi}, \clusters^{\pi}) = p(T^{\pi}) p(\clusters^{\pi} | T^{\pi})$. Further, we propose a canonical ordering of the subgraphs $\clusters^{\pi} = (C_1^{\pi}, \cdots, C_r^{\pi})$ in Section~\ref{sec:tree_generator}, which provides a sequential representation of the distribution:
\begin{equation}
p(T^{\pi}, \clusters^{\pi}) = p(T^{\pi}) \prod_i{p(C_{i}^{\pi} | T^{\pi}, C_{<i}^{\pi})}.
\end{equation}
As a result, graph generation is decomposed into two steps: (a)~generation of the tree $T^{\pi}$, and (b)~sequential generation of each subgraph $C_{i}^{\pi}$. In the following sections, we explain each step separately. 
Training is performed by likelihood maximization over a dataset of graphs and teacher-forcing is used between the steps. Because of teacher-forcing, each model can be trained independently.
After each training epoch, the nodes are reordered to achieve a general model which learns the complete distribution of the graphs and reduces overfitting to the limited training sequences. Technical details of the model are explained in Appendix~\ref{appendix:model_details}.

\subsection{Permutation Invariant Tree Generator}
\label{sec:tree_generator}
We propose a novel permutation invariant tree representation for tree generation, namely path length representation ($\mathsf{PLR}$).  Generating a $\mathsf{PLR}$ corresponds to a unique tree and vice versa. To obtain permutation invariance in $\mathsf{PLR}$, a permutation-invariant node ordering is required. For this, we introduce a canonical ordering over tree nodes inspired by the canonical naming proposed in the AHU algorithm for the isomorphism test on trees~\citep{aho1974design,campbell1991tree}. It is proven that two rooted trees are isomorphic \emph{iff} they have the same canonical name for their roots. Given this, we define a canonical ordering over the tree nodes by starting from the root and traversing the tree using depth-first search (DFS) while visiting the children of each node based on the dictionary order of their canonical names. The algorithm to calculate the canonical names and find the canonical root of a tree is explained in Appendix~\ref{sec:can-name}. Figure~\ref{fig:PLR_illustrate}(a) illustrates the canonical naming for an example tree.

\begin{algorithm}[t]

\SetAlgoLined
\For{ $u$ in $sortedChildren(v)$}{
    $\mathsf{DFS}\_\mathsf{PLR}(u, l+1)$\;
    $l=0$\;
}

$\mathsf{PLR}.append(l)$\;
$return$\;

 \caption{$\mathsf{DFS}\_\mathsf{PLR}(v,l)$}
 \label{alg_PLR}
\end{algorithm}

\begin{figure}[t!]
\vspace{-0.1in}
  \centering
   \includegraphics[width=78mm]{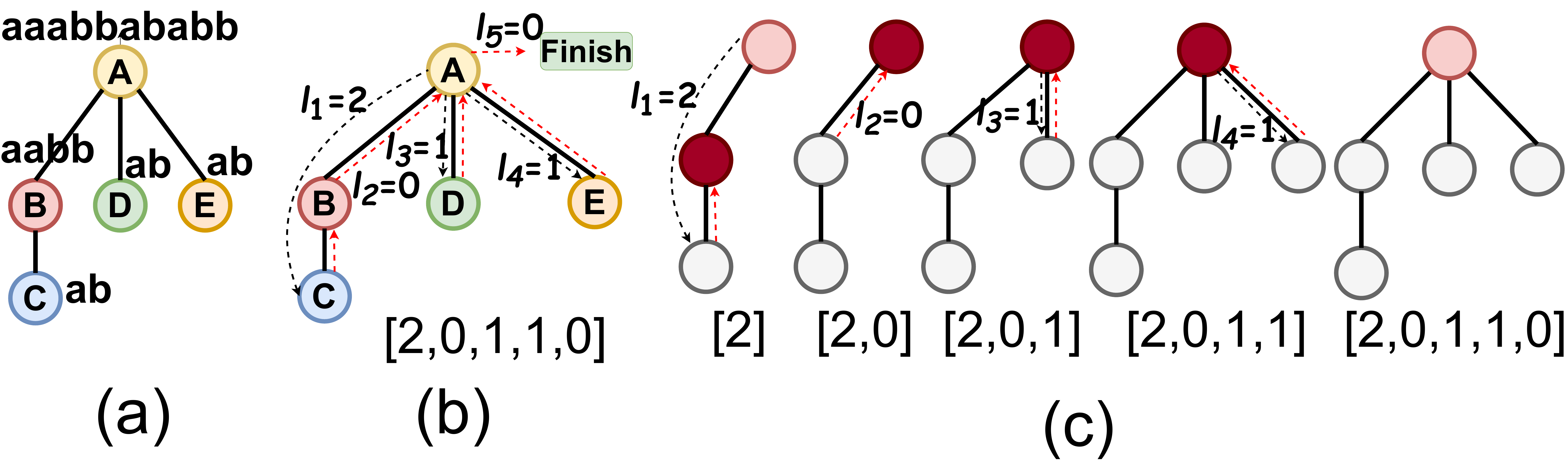}
  \caption{ (a) Assignment of the canonical names (shown above the nodes) to an example tree, which gives a canonical ordering of the tree nodes $[A, B, C, D, E]$. (b) $\mathsf{PLR}$ calculation for the example tree. Black dashed arrows show the forward paths in the DFS. The lengths of these paths are added to the $\mathsf{PLR}$. The Red dashed arrows show the backward paths. To represent the backward paths for the inner roots, a $0$ is added to the $\mathsf{PLR}$. (c) Step-by-step regeneration of the tree from the $\mathsf{PLR}$.
  }
  \label{fig:PLR_illustrate}
  \vspace{-0.1in}
  
\end{figure}

\subsubsection{Path Length Representation}

Given a tree $T$ and its $root$, the $\mathsf{PLR}$ of $T$ is defined as a sequence of non-negative integers $[\ell_1,\cdots,\ell_r]$ where $\ell_i$ is the length of the longest path from the current node
to the leaves in its subtree, excluding any visited edges in the previous $j<i$ steps. A recursive $\mathsf{PLR}$ algorithm is given in Algorithm~\ref{alg_PLR}, where $\mathsf{PLR}(T) = \mathsf{DFS}\_\mathsf{PLR}(root,0)$.
In Figure \ref{fig:PLR_illustrate}, the $\mathsf{PLR}$ of an example tree is illustrated.

\textbf{Theorem 1.} There is a bijection between the $\mathsf{PLR}$s of length $r$ and the space of the isomorphic classes of trees with $r$ nodes.\footnote{Proofs for all theorems and propositions are provided in Appendix~\ref{appendix:proofs}.}

According to Theorem 1, we can learn a distribution over the space of the trees by learning a distribution over its $\mathsf{PLR}$s. Note that not all sequences of non-negative integers are valid $\mathsf{PLR}$s. 
However, given $\ell_1, \ldots, \ell_{i-1}$, we can always find $(a_i, b_i)$ such that $a_i \leq \ell_i \leq b_i$ is necessary and sufficient for making a valid $\mathsf{PLR}$. The process of finding these boundaries is described in~Appendix~\ref{appendix:PLR-boundaries}.

\subsubsection{PLR Model of Tree Generation}
\label{sec:tree-gen-plr}
We propose to generate valid $\mathsf{PLR}$s by autoregressively generating the sequence $[\ell_1,\cdots,\ell_r]$. Given a prefix of a $\mathsf{PLR}$, a tree associated with that prefix and the next node where we want to extend the next path can be uniquely constructed (See Algorithm~\ref{alg:PLR_to_Tree} in Appendix~\ref{appendix:PLR-boundaries}). Let $T_{i-1}$ be the tree constructed from the prefix sequence $[\ell_1,\cdots,\ell_{i-1}]$, and $u_i$ be the node where the next path with length $\ell_i$ will be extended. Autoregressive generation of the $\mathsf{PLR}$ is modeled as
\begin{equation*}
\label{eq:tree_nll}
    p(\ell_1, \ell_2, \ldots , \ell_r) = \prod_i p(\ell_i | \ell_{j<i}) = \prod_i p(\ell_i | T_{i-1}, u_{i}),
\end{equation*}
where at each step, the goal is to model $\ell_{i}$ generation as $p(\ell_i | T_{i-1}, u_{i})$. However, to generate a valid $\mathsf{PLR}$, the $\ell_i$ value should be within a specified range $[a_i, b_i]$. This is because not necessarily any sequence of integers is a valid $\mathsf{PLR}$. In appendix~\ref{appendix:PLR-boundaries}, we show how we can calculate these boundaries for having a valid autoregressive $\mathsf{PLR}$ generation process.
To model $p(\ell_i | T_{i-1}, u_{i})$, first a standard tree encoder with message passing is used to encode the tree and obtain the node representations $H_T[\cdot]$ for all the nodes in $T_i$ (See Section~\ref{sec:treeDetails}). Next, $\ell_i$ generation is modeled with a Multinouli distribution $p(\ell_i = x | T_{i-1}, u_{i}) = p_x$ for $a_i \leq x \leq b_i$, where $p_x$ is defined as
\begin{eqnarray}
    (p_{a_i},\ldots,p_{b_i}) &=& f_{\theta}(H_T[root],H_T[u_i],M).
\end{eqnarray}
Here, $M$ is a binary mask vector representing the boundaries of $\ell_{i}$, and $f_{\theta}$ is a parametric model implemented as an MLP with an output normalizer (\emph{e.g.} softmax) to make the normalized probabilities for the Multinoulli distribution. 
In other words, the root encoding $H_T[root]$ (as a representation of the entire $T_i$), the current node encoding $H_T[u_i]$, and the mask vector are concatenated as the input of a neural network which outputs the probabilities to generate $\ell_{i}$. For handling the variable length of the output we use masking. We mask the output to give a positive probability to all values between $a_i$ and $b_i$, and zero probability to the other values.

\textbf{Theorem 2.} Each valid $\mathsf{PLR}$ can be generated by the proposed model, and every $\mathsf{PLR}$ generated using this model is valid.

\begin{figure}[t!]
  \centering
  \includegraphics[width = 7.5cm]{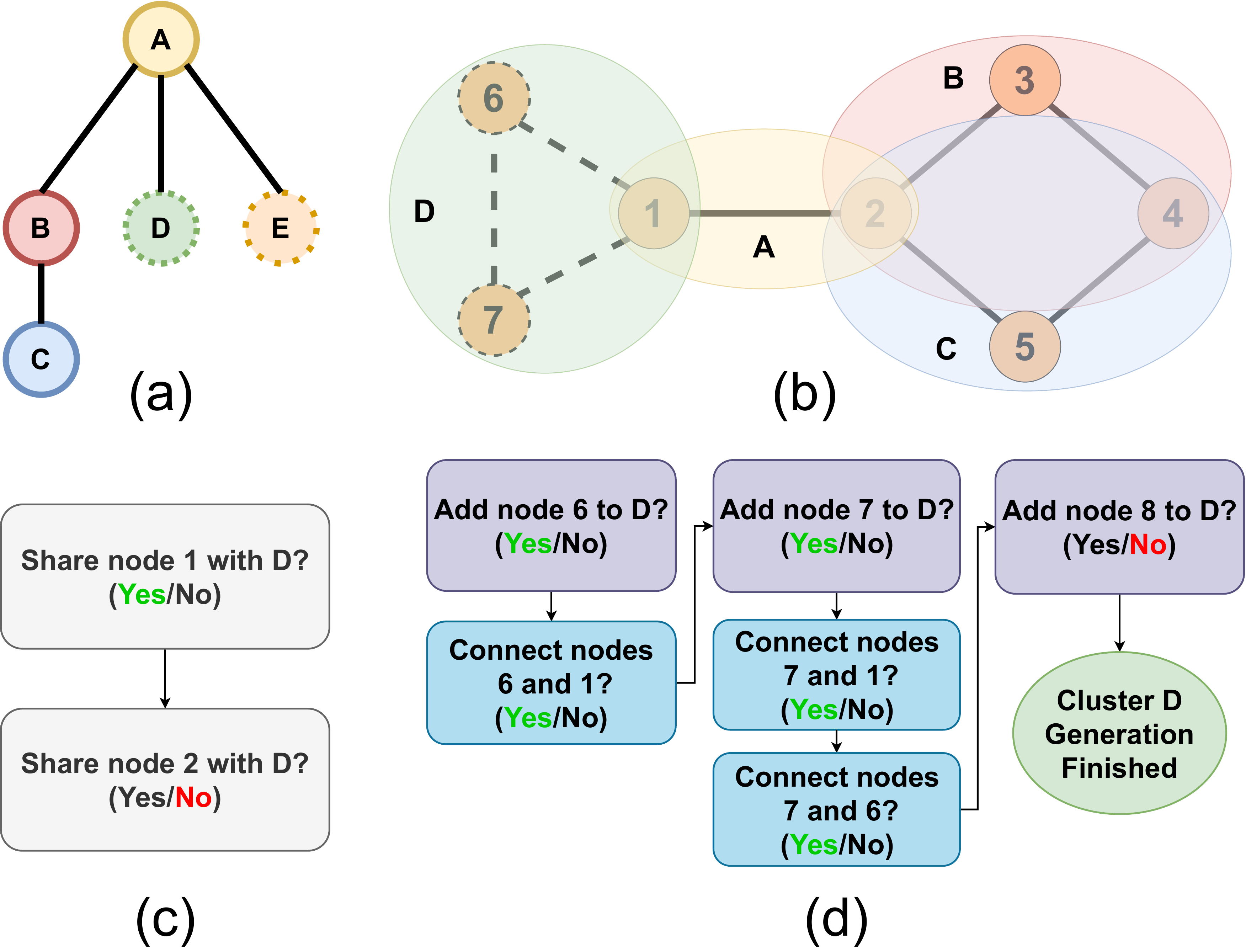}
  \caption{Steps for generating the subgraph "D". a) The tree structure for the tree decomposition. The supernodes whose corresponding subgraph is not created yet are shown with dashed lines. b) The nodes and connections generated so far, and the dashed nodes and connections to be generated in this step. c) Steps for autoregressively deciding about sharing the nodes from the parent subgraph "A". Green "Yes" or Red "No" highlights the decision made. d) Steps for adding new nodes to make a complete subgraph.}
\label{fig:sub-graph-generation}
\vspace{-0.1in}
\end{figure}

\subsubsection{Tree Encoder}
\label{sec:treeDetails}
For encoding trees, we use a similar model to JT-VAE \citep{jin2018junction} which uses a two-phase message passing process: one from the leaves to the root and the other from the root to the leaves. 
In the process we 
learn the initial encoding of the nodes, so that before message propagation tree node $i$ is initialized with the encoding $E_i$. 
Gathering the messages is done using a Gated Recurrent Unit (GRU)~\citep{chung2014empirical}. 
All $m_{ij}$s are initially set to null, followed by a two-phase procedure as detailed here:
$$
m_{i j} = GRU( E_i, \{ m_{k,i}| k \in  N(i) - j \text{ and } m_{k,i} \neq null \}).
$$
The final representation of each supernode is calculated as:
$$
H_T[i]=ReLU(W (E_i || \sum_{k \in N(i)}  m_{k i})),
$$
where $||$ is the concatenation operation. After finishing the process, we use the encoding of the root as the encoding of the tree.

\subsection{Generating Subgraphs}
\label{sec:gen_cluster}

Once the tree is generated, it is traversed following the canonical order to generate each subgraph via two steps: (1) finding the nodes shared by the parent subgraph (except for the tree root which has no parent), and (2) adding new nodes and edges within the subgraph. These steps are used alternately to create new subgraphs. Figure~\ref{fig:sub-graph-generation} illustrates an example of this process.

\textbf{Node Sharing:} The goal is to find the nodes from the parent subgraph 
to be shared with the current subgraph (corresponding to the tree supernode $u$). This can be done with an autoregressive model to generate a binary sequence $[s_1, ..., s_m]$, where $s_i$ denotes whether the node $v^p_i$ from the parent subgraph is shared with the current subgraph. To this end, first a graph attention network (GAT)~\citep{velivckovic2017graph,liao2019efficient} is used to encode the partially generated graph and construct the node representations $H_G^S[v^p_i]$\footnote{
In subscripts, $G$ and $T$ stand for graph and tree. In superscripts, $S$ and $A$ stand for node sharing and adding.} for the nodes in the parent subgraph. A tree encoder (as explained in Section~\ref{sec:tree-gen-plr}) is also used to obtain the tree node encodings $H_T^S$ for the tree supernodes. Given these encodings, the probability of sharing the node $v^p_i$ is defined as 
\begin{equation*}
p(s_i=1|s_{<i}) = f_{\phi}(H_G^S[v^p_i], H_T^S[root], H_T^S[u], s_{<i}),
\end{equation*}
where $f_{\phi}$ is a parametric function implemented as an MLP with a sigmoid output activation. 
To fix the size of $s_{<i}$, we can mask a longer fixed-sized vector.  The length of this vector can be obtained based on the maximum size of the subgraph nodes.

\textbf{Adding New Nodes:} This stage comprises adding new nodes and deciding about the connections between the new nodes and the existing nodes inside the subgraph. This process is very similar to autoregressive models such as GraphRNN or GRAN. However, since the node and edge additions are intra-cluster, the decision space can be significantly smaller than the total nodes of the entire graph. Node adding is modeled as deciding if a new node should be added or not. This is done one-by-one until the model decides no new node is added.
At each step, first, a GAT model is used to find the node encodings $H_G^A[v]$ for the nodes of the partially generated graph. Accordingly, $\tilde{H}_G = \sum_{i=1}^n H_G^A[v_i]$ is defined as the encoding of the entire partially generated graph. Given this and the tree supernode encodings $H_T^A[.]$ (similar to node sharing), the probability of adding a new node $v_j$ is modeled as
\begin{equation}
p(v_{j}) = f_{\psi}(\tilde{H}_G, H_T^A[root], H_T^A[u]),
\end{equation}
where $f_{\psi}$ is a parametric model implemented as an MLP with a sigmoid output activation.
Likewise, the probability of adding an edge between an existing subgraph node $v_i$ and the newly generated node $v_j$ is modeled as
\begin{multline}
p(\boldsymbol{A}_{v_j, v_i} = 1 | \boldsymbol{A}_{v_j, v_{<i}}) = \\ f_{\gamma}(\tilde{H}_G, H_G^A[v_i], H_T^A[root], H_T^A[u],  \boldsymbol{A}_{v_j, v_{<i}}),
\end{multline}
where $\boldsymbol{A}$ represents the adjacency matrix and $f_{\gamma}$ is a parametric model with a similar structure as $f_{\psi}$. 
Similarly, to fix the size of $\boldsymbol{A}_{v_j, v_{<i}}$, we use the same masking technique as in node sharing with the same approach to get the vector length.

\section{COMPLEXITY ANALYSIS}
\label{complexity}
In this section, we discuss the complexity of our model during generation and also provide a framework for analysing the hypothesis space complexity and give some experimental results on how our model reduces the hypothesis space.

\vspace{-0.1in}

\subsection{Graph Generation Time Complexity}
\label{sec:generation-complexity}
\vspace{-0.08in}
The proposed model's time complexity is analyzed with respect to the number of steps for generating a graph. 
Particularly, generation steps can be broken down into two phases: 1) Generating the tree: Proposition 3 shows how a minimal tree decomposition can restrict the number of tree nodes, and consequently, the tree generation can be done in $O(n)$ steps. 
This proposition is important because 
the tree is a simpler structure and required time for its generation is negligible compared to the graph structure. This is true in both the tree decomposition and tree generation processes.
2) Node sharing and node adding: Theorem 4 shows how the total number of steps in this phase scales with the width of the tree decomposition.

Proposition 5 gives an upper bound for the treewidth of a graph.
This upper bound is the result of the Breadth-First Search (BFS) algorithm for tree decomposition. BFS has been previously used for reducing the number of steps in GraphRNN \citep{you2018graph}. Here, we show that BFS is a special case of tree decomposition. Hence, it can be used for tree decomposition and leads to a similar reduction as in GraphRNN; there exist better tree decompositions that can further reduce the number of steps.
Proofs are provided in Appendix~\ref{appendix:proofs}.

\textbf{Proposition 3.} Given a graph $G$ with $n$ nodes and a minimal tree decomposition of it, e.g., $\mathcal{T} = (T,C)$, with width $k$, then $r \leq n-k+1$, where $r$ is the number of nodes in $T$.

\textbf{Theorem 4.} Generating a graph with $n$ nodes using a tree decomposition with width $k$ 
has worst-case time complexity of $O(nk)$ with respect to the decision steps.

\textbf{Proposition 5.} For an arbitrary ordering of the nodes $(v_1, \cdots, v_n)$ of a connected graph $G$,
$$tw(G) \leq 2 \times \max_{1 \leq d \leq \operatorname{diam}(G)}
\Big| \{v_{i} | \operatorname{dist}(v_{i}, v_{1})=d\}\Big|$$ where $diam(G)$ is the diameter of the graph $G$ and $dist$ is the distance function on the $G$. 

\subsection{Hypothesis Space Complexity in Training} 
\label{sec:learning-complexity}
\vspace{-0.08in}

\begin{figure}[!t]
  \centering
  \includegraphics[width = 7.5cm]{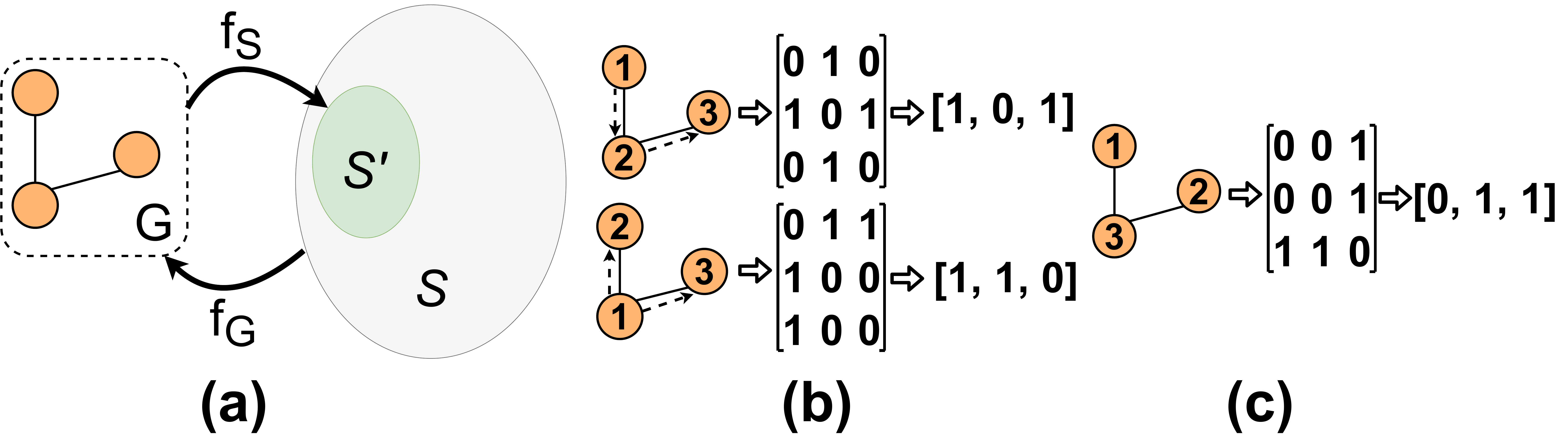}
  \caption{ a) $\mathcal{S'}$ shows all the sequences that can be generated from $G$ using the mapping $f_S(\cdot)$. $\mathcal{S}$ shows all possible sequences that can be mapped to $G$ using $f_G(\cdot)$. b) Two sequences created from an example graph using BFS ordering and sequencing the adjacency matrix using the elements on the upper triangle. Numbers on the nodes indicate their corresponding order in the adjacency matrix, so node 1 is the first node for the BFS algorithm. Arrows show the direction of the edges in the BFS tree. These sequences create $\mathcal{S'}$. c) A sequence which cannot be generated by BFS (\emph{i.e.} it is not in $\mathcal{S'}$). However, the sequence is mapped to the same graph.}
  \label{fig:gen-framework}
  \vspace{-0.1in}
\end{figure}

As explained in Section~\ref{sec:model-formulation}, sequential graph generation is formulated as a mapping $f_S$ from graphs to sequences. Let $G$ be a fixed graph. By considering different node orderings $\pi \in \Pi$, $f_S(G, \pi)$ will map $G$ to a set of sequences $\mathcal{S'}$. 
On the other hand, the mapping $f_G$ exists which maps sequences to $G$. 
As shown in Figure~\ref{fig:gen-framework},
$\mathcal{S} = \{S | f_G(S)=G\}$ is the domain of $f_G$, 
and $\mathcal{S'} = \{S | f_S(G)=S\}$ is the range of $f_S$, 
while $\mathcal{S} \neq \mathcal{S'}$.

When formulating graph generation as learning the graph distribution,
\begin{equation}
\label{eq:P(G)}
    p_{model}(G) = \sum_{S \in \mathcal{S}} p_{model}(S),
\end{equation}
which means the likelihood is marginalized over all possible sequences in $\mathcal{S}$.
However, in the sequential graph generation models, sequences are sampled from $\mathcal{S'}$ during training.
In this way, the model is trained to give high probability to the sequences in $\mathcal{S'}$. On the other hand, because of the constraints on the marginal likelihood of the density functions, the model learns to give lower probability to the sequences in $\mathcal{S}-\mathcal{S'}$. For a perfect model,
\begin{multline}
     \forall s \in \mathcal{S}-\mathcal{S'}: p_{model}(s) = 0  \\
    \Rightarrow  p_{model}(G) = \sum_{s \in \mathcal{S'}} p_{model}(s),
\end{multline}
which can only be approximated for an imperfect model as 
$p_{model}(G) \approx \sum_{S \in \mathcal{S'}} p_{model}(S)$.
Prior works have attempted to use standard algorithms 
to decrease the size of $\mathcal{S'}$ (\emph{e.g.}, GraphRNN uses BFS)~\citep{you2018graph,liu2019graph}.

\begin{figure}[t!]
\centering
\subfloat[]{\includegraphics[width = 1.68in]{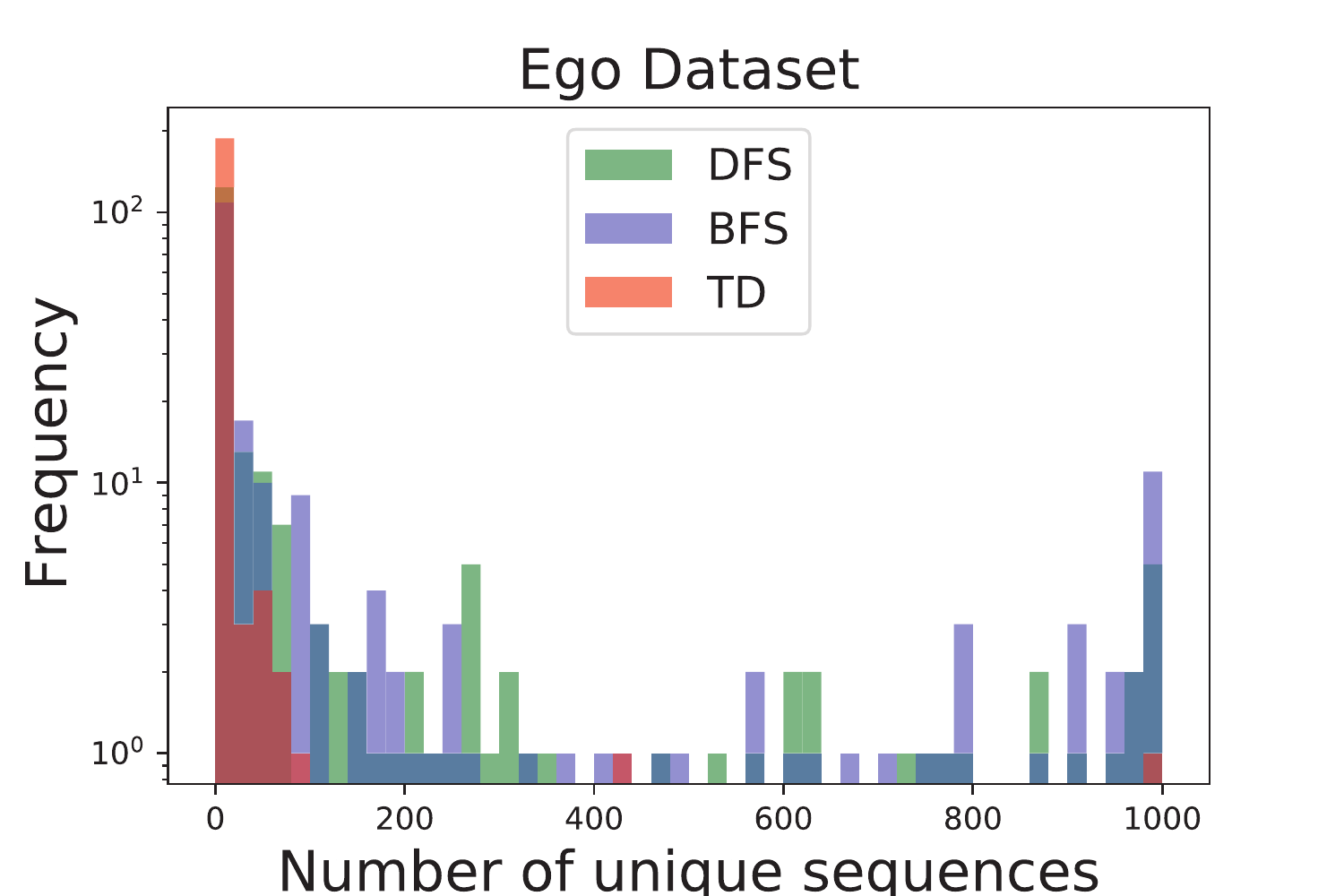}}
\subfloat[]{\includegraphics[width = 1.68in]{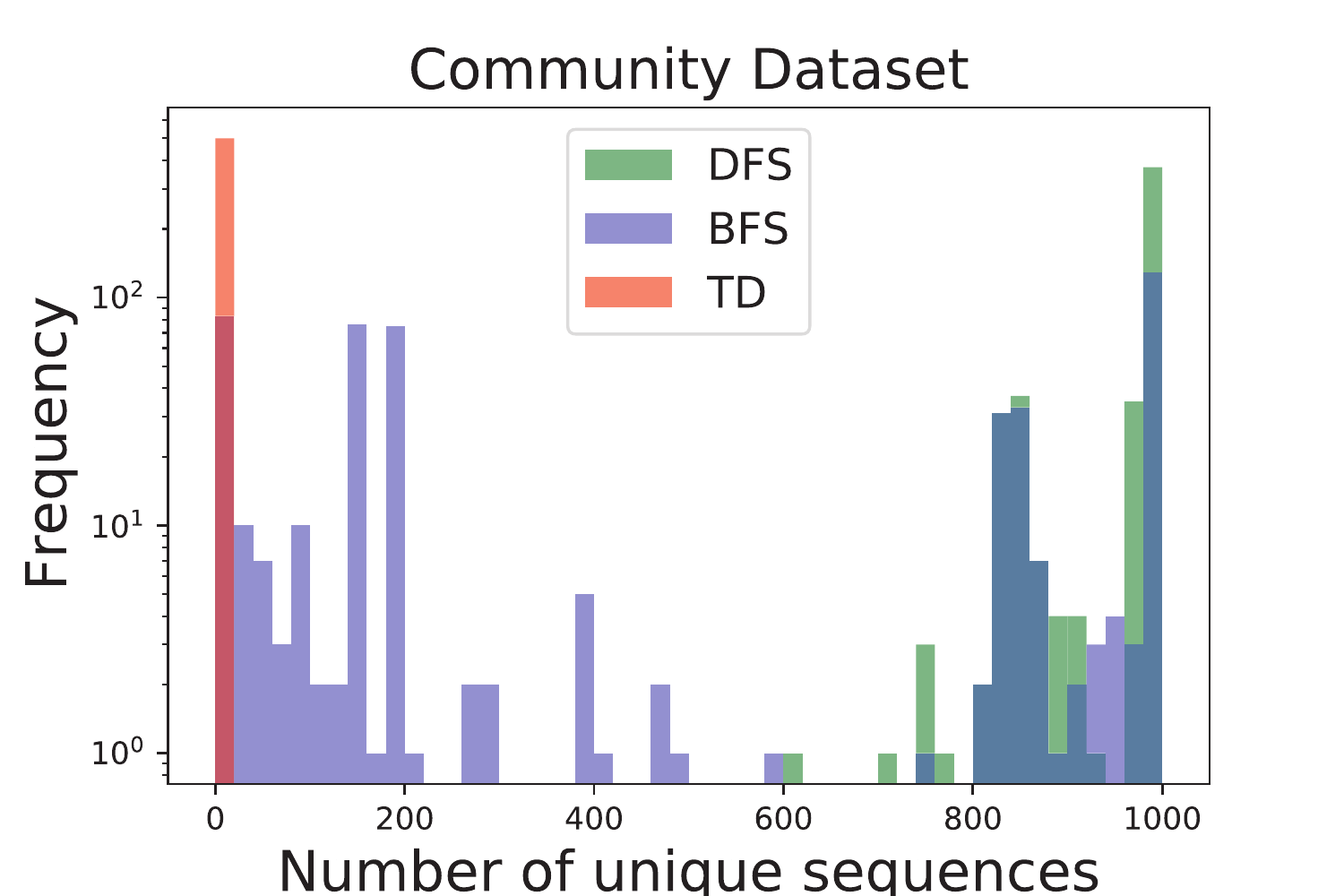}}
\caption{Statistical comparison of the number of unique sequences generated from $1000$ random initial node permutations for each graph in two datasets using three methods: BFS, DFS, and tree decomposition (TD). The results show that the TD is superior to have a significant reduction in the size of $\mathcal{S'}$ compared to BFS and DFS. }
\label{fig:num_orders}
\vspace{-0.1in}
\end{figure}

Based on the above, we introduce some properties of an efficient sequential graph generation model:

    1. $\mathcal{S'}$ is as small as possible.  This helps integrate over $\mathcal{S'}$ to find $p(G)$ more efficiently. Also, as $\mathcal{S'}$ gets smaller, learning becomes simpler since the hypothesis space is smaller. In Figure~\ref{fig:num_orders}, we have experimentally analyzed the size of $\mathcal{S'}$ on two datasets of graphs with maximum $20$ nodes. It can be observed that our tree decomposition gives a better mapping to significantly reduce $\mathcal{S'}$. Thus, $p(G)$ can be calculated more efficiently. In fact, by marginalizing over the small set of sequences in $\mathcal{S'}$, our approach can become comparable to permutation invariant models for small graphs.
    
    2. $\mathcal{S}-\mathcal{S'}$ is as small as possible. In the ideal case, we have $\mathcal{S} = \mathcal{S'}$. In this way, training is focused on the right hypothesis space without trying to implicitly decrease the probability of the model on $\mathcal{S}-\mathcal{S'}$. Given this, pruning $\mathcal{S}-\mathcal{S'}$ is a valid strategy for improving model efficiency. We use the Fill-in algorithm~\citep{bodlaender2010treewidth} for tree decomposition. However, given an arbitrary graph, the space of tree decompositions (\emph{i.e.}, $\mathcal{S}$) is infinite as each subgraph can repeat infinitely. To handle this problem, we use the minimal tree decomposition introduced in Appendix~\ref{appendix:td-alg}. Furthermore, during training and for the subgraph node adding, the algorithm is modified such that the first decision of the node adding is always "Yes". This ensures every possible tree decomposition is minimal and considerably prunes $\mathcal{S}$. Other pruning strategies can potentially be pursued.
    
    3. The model should effectively learn to assign higher probabilities to the sequences in $\mathcal{S'}$ and lower probabilities to the sequences in $\mathcal{S}-\mathcal{S'}$. In other words, the model should have enough capacity to learn the difference between the sequences in $\mathcal{S'}$ and $\mathcal{S}-\mathcal{S'}$. As the integration of the probabilities is fixed to be $1.0$, this is equivalent to maximizing the likelihood on $\mathcal{S'}$. 

\section{EXPERIMENTS \& RESULTS}
\label{sec:experiemnts}

In this section, we first discuss the shortcomings of the statistical metrics, how they fail to fairly assess the generalizability of graph generation models, and suggest a likelihood measure as a more suitable substitute. 
Finally, we show how the proposed model outperforms other approaches across metrics and datasets in standard vs. overfitting settings.

\vspace{-0.1in}
\subsection{Performance Measures and Overfitting}
\vspace{-0.1in}
The standard performance metrics of graph generation models compare the statistics of the generated graphs with those of a held-out test set. 
The commonly used statistics include: 
(1) degree distributions, 
(2) clustering coefficient distributions, and 
(3) distributions over occurrences of orbits with four nodes.  Accordingly, Maximum Mean Discrepancy (MMD) with a Gaussian Earth Mover's Distance (EMD) kernel or total variation (TV) are used to estimate the distances between distributions.
As is common, the training and test sets are assumed to come from the same distributions; hence, these statistics are expected to match across the two datasets, particularly in synthetic data.
Therefore, we argue that these metrics are problematic as they over-estimate the performance of an overfitted high-variance model, which merely reproduces the training set.
However, the likelihood of generating the unseen test graphs using such a model could be very small due to its lack of generalizability.

We propose using the gap between \emph{expected value of negative log-likelihood (NLL)} across node permutations, between the train and test data, as a measure of generalization.
Expected NLL was also used in GraphRNN~\citep{you2018graph}, yet on small datasets as the number of permutations grows super-exponentially with respect to the number of nodes.
In our proposed method, the number of unique sequences is reduced using tree decomposition; hence, NLL can be more efficiently estimated.
It is worth highlighting that expected NLL is not a perfect metric as it over-estimates performance of models with smaller numbers of possible sequences; however, a smaller hypothesis space is deemed a desirable quality in a generative model.
Moreover, when comparing performance of the same model across datasets or comparing models with similar sequencing strategy, expected NLL is an objectively good metric.
 
 \vspace{-0.1in}
\subsection{Datasets}
\label{sec:datasets}
\vspace{-0.1in}
The proposed model was empirically evaluated across the following datasets~\citep{you2018graph}.
(1) Community-small with 500 two-community graphs with $12 \leq |V| \leq 20$. (2) Ego-small with 200 ego graphs extracted from the Citeseer network with $4 \leq |V| \leq 18$.
(3) community with 500 two-community graphs with $60 \leq |V| \leq 160$, probability of inside community connections are $0.7$, and $0.05|V|$ edges uniformly selected between communities. (4) Ego with 299 3-hop ego graphs extracted from the Citeseer network with $50 \leq |V| \leq 100$.
In all experiments, the dataset was split by 70/10/20 ratios for train/validation/test accordingly. 
\vspace{-0.1in}
\subsection{Statistical Evaluation}
\label{sec:stat-based-eval}
\vspace{-0.1in}
Table~\ref{table:overfit} compares the proposed model with the state-of-the-art permutation-invariant and sequential models on two graph generation datasets using statistical metrics.
In this table, the TD-GEN model is trained with the standard training settings using early-stopping with respect to the NLL metric.
On the contrary, the TD-GEN* is allowed to overfit on the training data. 
As a result, the TD-GEN model has a higher (more balanced) train to test NLL ratio as demonstrated in Table~\ref{table:nll}. 

As shown in Table~\ref{table:overfit}, with both standard and overfitting settings, our models demonstrate the best performance.
Yet, the overfitted model (TD-GEN*) maintains a bigger superiority margin across the statistical metrics. 
This is an empirical evidence of these metrics' shortcomings as the less-generalizable model is shown to obtain better results. Table~\ref{tab:lobster-res} in Appendix~\ref{sec:lobster} does a similar comparison for Lobster trees generation.

\begin{table*}[t!]
\centering

\caption{Graph generator models' performance with respect to statistical measures. TD-GEN is the proposed model with early stopping with respect to the validation NLL. TD-GEN* is the same model overfitted to the training data. The final row shows the distance of the train dataset from the test dataset.
For statistical distances, the number of generated graphs is equal to the size of the test dataset.
}
 \scalebox{0.75}{
\begin{tabular}{@{}l|lll|lll@{}}
\toprule
\multirow{2}{*}{Model} & \multicolumn{3}{c|}{Community-small} & \multicolumn{3}{c|}{Ego-small} \\ \cmidrule(l){2-7} 
 & Deg. & Clus. & Orbit & Deg. & Clus. & Orbit \\ \midrule
GraphVAE~\citep{simonovsky2018graphvae} & $0.350$ & $0.980$ & $0.540$ & $0.130$ & $0.170$ & $0.050$ \\
DeepGMG~\citep{li2018learning} & $0.220$ & $0.950$ & $0.400$ & $0.040$ & $0.100$ & $0.020$ \\
GNF~\citep{liu2019graph} & $0.200$ & $0.200$ & $0.110$ & $0.030$ & $0.100$ & $\mathbf{0.001}$ \\
EDP-GNN~\citep{niu2020permutation} & $0.053$ & $0.144$ & $0.026$ & $0.052$ & $0.093$ & $0.007$ \\
GraphAF~\citep{shi2020graphaf} & $0.180$ & $0.200$ & $0.020$ & $0.030$ & $0.110$ & $\mathbf{0.001}$ \\
GraphRNN~\citep{you2018graph} & $0.080$ & $0.120$ & $0.040$ & $0.090$ & $0.220$ & $0.003$ \\
GRAN~\citep{liao2019efficient} & $0.152$ & $0.142$ & $0.032$ & $0.030$ & $0.029$ & $0.014$ \\
TD-GEN [ours] & $0.024$ & $0.034$ & $0.005$ & $0.014$ & $0.077$ & $0.005$ \\
TD-GEN* [ours] & $\mathbf{0.010}$ & $\mathbf{0.008}$ & $\mathbf{0.002}$ & $\mathbf{0.006}$ & $\mathbf{0.015}$ & $\mathbf{0.001}$ \\ \midrule
Train Dataset & $0.002$ & $0.0005$ & $0.0018$ & $0.009$ & $0.019$ & $0.007$ \\ \bottomrule
\end{tabular}
 }
\label{table:overfit}

\end{table*}

\subsection{Likelihood Based Evaluation}
In this section, our model is compared with the likelihood-based sequential graph generation models using the NLL criterion on both training and test datasets (Table~\ref{table:nll}). 
Our proposed models, in both overfitting and standard settings, achieve the best results across all datasets. However, by overfitting, the model loses some performance on the test NLL compared to the standard TD-GEN.
Moreover, comparing with the other models (\emph{e.g.}, GRAN and GraphRNN), TD-GEN has a smaller gap between the train and test NLL, which demonstrates its generalizability.

\begin{table*}[t]

\centering
\setlength{\tabcolsep}{2.5pt}
\vskip 0.2in
\caption{
Graph generator models' performance with respect to the NLL metric on the train and test datasets. We used as many different permutations until we noticed no change in the estimated NLL.
}
 \scalebox{0.7}{
\begin{tabular}{@{}l|cc|cc|cc|cc@{}}
\toprule
\multirow{2}{*}{Model} & \multicolumn{2}{c|}{\begin{tabular}[c]{@{}c@{}}Community-small\end{tabular}} & \multicolumn{2}{c|}{\begin{tabular}[c]{@{}c@{}}Ego-small\end{tabular}} & \multicolumn{2}{c|}{Community} & \multicolumn{2}{c}{\begin{tabular}[c]{@{}c@{}}Ego\end{tabular}} \\ \cmidrule(l){2-9} 
 & Train & Test & Train & Test & Train & Test & Train & Test \\ \midrule
\small{GraphRNN-S}~\citep{you2018graph} & $31.24$ & $35.94$ & $8.51$ & $9.88$ & $2019$ & $2041$ & $248.86$ & $253.6$  \\
\small{GraphRNN}~\citep{you2018graph} & $28.95$ & $35.10$ & $9.05$ & $10.61$ & $1835$ & $1968$ & $237.05$ & $243.34$  \\
\small{GRAN}~\citep{liao2019efficient} & $\mathbf{9.42}$ & $23.04$ & $7.86$ & $43.24$ & $3752$ & $3558$ & $342.06$ & $735.75$  \\
\small{TD-GEN} [ours] & $14.98$ & $\mathbf{17.62}$ & $7.73$ & $\mathbf{6.36}$ & $1742$ & $\mathbf{1803}$ & $232.54$ & $\mathbf{234.69}$  \\
\small{TD-GEN*} [ours] & $12.81$ & $18.63$ & $\mathbf{5.54}$ & $7.03$ & $\mathbf{1708}$ & $1842$ & $\mathbf{176.04}$ & $317.77$  \\ \bottomrule
\end{tabular}
 }
\label{table:nll}

\end{table*}

\subsection{Tree Generation Compared with JT-VAE}
In this section, we evaluate our proposed tree generator (Tree-Gen) by comparing with JT-VAE~\citep{jin2018junction} for tree generation. JT-VAE has been proposed for molecule generation and uses a vocabulary of predetermined structures (\emph{e.g.},  rings, bonds, and atoms) to map molecules to trees. Thus, it cannot be used for general graph generation. But, it introduces a tree-based variational autoencoder for general tree generation. Here, we compare our permutation invariant tree generator with the variational tree generator in JT-VAE for learning the tree structures derived from tree decomposition. The results of the comparison are shown in Table~\ref{table:JT-VAE}. The likelihood values for JT-VAE have been approximated by the variational lower bound.

It can be observed in Table~\ref{table:JT-VAE} as the trees grow larger (\emph{e.g.} from the Ego-small and Community-small datasets to the Ego and Community datasets), our proposed model takes a larger advantage over JT-VAE.

\begin{table}[ht!]
\centering
\setlength{\tabcolsep}{2.5pt}
\vskip 0.2in
\caption{Comparison between our tree generator and the JT-VAE tree generator using both statistical and likelihood based metrics. For the statistical metrics, the number of generated graphs is equal to the size of the test dataset.}
\label{table:JT-VAE}
\small
\scalebox{0.75}{
\begin{tabular}{@{}lllllllll@{}}
\toprule
Dataset                                               & Model                         & Deg.         & Clus. & Orbit         & Spec.   & Train & Test \\ \midrule
\multicolumn{1}{l|}{\multirow{2}{*}{Community-small}} & \multicolumn{1}{l|}{JT-VAE}   & $4.19e^{-2}$ & $0.0$ & $1.09e^{-3}$  & $2.56e^{-1}$      & $3.28$    & $4.91$   \\
\multicolumn{1}{l|}{}                                 & \multicolumn{1}{l|}{Tree-GEN} & $2.04e^{-3}$ & $0.0$ & $2.26e^{-4}$  & $1.07e^{-1}$      & $2.48$    & $2.25$   \\ \midrule
\multicolumn{1}{l|}{\multirow{2}{*}{Ego-small}}       & \multicolumn{1}{l|}{JT-VAE}   & $9.04e^{-3}$ & $0.0$ & $7.56e^{-3}$  & $6.27e^{-3}$      & $2.87$    & $5.81$   \\
\multicolumn{1}{l|}{}                                 & \multicolumn{1}{l|}{Tree-GEN} & $7.13e^{-3}$ & $0.0$ & $1.55e^{-1}$  & $9.65e^{-2}$      & $2.46$    & $1.60$   \\ \midrule
\multicolumn{1}{l|}{\multirow{2}{*}{Community}}       & \multicolumn{1}{l|}{JT-VAE}   & $1.40e^{-1}$ & $0.0$ & $8.27e^{-3}$  & $2.74e^{-1}$      & $12.18$   & $14.97$  \\
\multicolumn{1}{l|}{}                                 & \multicolumn{1}{l|}{Tree-GEN} & $1.18e^{-3}$ & $0.0$ & $2.35e^{-5}$  & $2.26e^{-3}$  & $10.52$   & $11.48$  \\ \midrule
\multicolumn{1}{l|}{\multirow{2}{*}{Ego}}             & \multicolumn{1}{l|}{JT-VAE}   & $1.5e^{-2}$  & $0.0$ & $1.26$        & $9.27e^{-2}$  & $49.82$   & $56.39$  \\
\multicolumn{1}{l|}{}                                 & \multicolumn{1}{l|}{Tree-GEN} & $1.40e^{-3}$ & $0.0$ & $1.88e^{-2}$  & $5.57e^{-3}$ & $41.30$   & $42.70$  \\ \bottomrule
\end{tabular}
}
\end{table}

\subsection{Ablation Studies}
The results of testing the impact of different components of the model are reported in Table~\ref{table:ablation}. 
First, we have removed tree decomposition (w/o TD) and generated the whole graph using the cluster generator. 
The results show that tree decomposition is a crucial part of the method.

In the second experiment, we used tree decomposition but did not include the tree features in node sharing and node adding stages. 
As demonstrated (Table~\ref{table:ablation}, w/o tree features), tree features are essential in the model's performance.

For the third part, we propose a simpler version of the method, namely TD-GEN-S, inspired by the GraphRNN-S. In this method, we do not use the masked results of the previous nodes and edges.
Accordingly, all the sharing in a cluster and all the nodes' connections within a cluster can be determined in parallel. 
As shown in the results, TD-GEN-S performs worse on the Community-small data while performs better in the Ego-small dataset.
This is justified given that the GraphRNN-S had shown to be slightly better than GraphRNN in the Ego-small data~\citep{you2018graph}.
But, in general, and considering the average performance on the two datasets, the results support that having the masked results of the previous nodes and edges can help the learning.

\begin{table}[h!]
\caption{Ablation study of the proposed model with respect to the NLL metric on the train and test datasets.}
\centering
\label{table:ablation}
\scalebox{0.7}{
\begin{tabular}{@{}l|ll|ll@{}}
\toprule
\multirow{2}{*}{Model} & \multicolumn{2}{c|}{Community-small} & \multicolumn{2}{c}{Ego-small} \\ \cmidrule(l){2-5} 
 & Train & Test & Train & Test \\ \midrule 
w/o TD
& $35.22$ & $36.40$ & $9.59$ & $7.66$ \\
w/o tree features & $18.29$ & $25.07$ & $8.34$ & $6.40$ \\
TD-GEN-S & $16.66$ & $22.78$ & $7.95$ & $6.20$ \\
TD-GEN & $14.98$ & $17.62$ & $7.73$ & $6.36$ \\
\bottomrule
\end{tabular}
}
\end{table}

\section{CONCLUSION}
In this paper, we presented TD-GEN, a graph generation approach based on tree decomposition.  We contribute a novel
permutation invariant model for generating trees based on a
canonical ordering for nodes and path length representation.  The tree decomposition forms the backbone of
the generated graph; a subsequent subgraph generation produces a graph respecting the backbone tree structure.

We proved that tree decomposition has essential advantages, including
significantly reducing the number of edge predictions required for
generation and greatly reducing the number of node permutations required to be considered by the model. These advantages are borne out in empirical studies over a variety of datasets. We also showed that statistics-based measures, commonly used to compare graph generation models, can be misleading and may not represent the performance of a model for learning the complete data distribution.

Hence TD-GEN is efficient for graphs with relatively small treewidth. For graphs with larger treewidth, time complexity can be as bad as the time required for generating the entire adjacency matrix. On the other hand, finding an optimal tree decomposition is an NP-hard problem. Optimal algorithms for tree decomposition for different kinds of graphs can be further explored in this direction. Also, the characteristics of these algorithms can help to improve the model from many perspectives. For example, as explained in Section~\ref{sec:learning-complexity}, the number of different tree decompositions and ratio  $\frac{\mathcal{S'}}{\mathcal{S}}$ can both be decreased using a minimal decomposition algorithm.

\newpage
\bibliography{bibliography.bib}

\begin{thebibliography}{}

\bibitem[Aho and Hopcroft, 1974]{aho1974design}
Aho, A.~V. and Hopcroft, J.~E. (1974).
\newblock {\em The design and analysis of computer algorithms}.
\newblock Pearson Education India.

\bibitem[Bodlaender and Koster, 2010]{bodlaender2010treewidth}
Bodlaender, H.~L. and Koster, A.~M. (2010).
\newblock Treewidth computations i. upper bounds.
\newblock {\em Information and Computation}, 208(3):259--275.

\bibitem[Bojchevski et~al., 2018]{bojchevski2018netgan}
Bojchevski, A., Shchur, O., Z{\"u}gner, D., and G{\"u}nnemann, S. (2018).
\newblock Netgan: Generating graphs via random walks.
\newblock {\em arXiv preprint arXiv:1803.00816}.

\bibitem[Campbell and Radford, 1991]{campbell1991tree}
Campbell, D.~M. and Radford, D. (1991).
\newblock Tree isomorphism algorithms: Speed vs. clarity.
\newblock {\em Mathematics Magazine}, 64(4):252--261.

\bibitem[Chen et~al., 2021]{chen2021order}
Chen, X., Han, X., Hu, J., Ruiz, F.~J., and Liu, L. (2021).
\newblock Order matters: Probabilistic modeling of node sequence for graph
  generation.
\newblock {\em arXiv preprint arXiv:2106.06189}.

\bibitem[Chung et~al., 2014]{chung2014empirical}
Chung, J., Gulcehre, C., Cho, K., and Bengio, Y. (2014).
\newblock Empirical evaluation of gated recurrent neural networks on sequence
  modeling.
\newblock {\em arXiv preprint arXiv:1412.3555}.

\bibitem[Dai et~al., 2020]{dai2020scalable}
Dai, H., Nazi, A., Li, Y., Dai, B., and Schuurmans, D. (2020).
\newblock Scalable deep generative modeling for sparse graphs.
\newblock {\em arXiv preprint arXiv:2006.15502}.

\bibitem[Goodfellow et~al., 2014]{goodfellow2014generative}
Goodfellow, I., Pouget-Abadie, J., Mirza, M., Xu, B., Warde-Farley, D., Ozair,
  S., Courville, A., and Bengio, Y. (2014).
\newblock Generative adversarial nets.
\newblock In {\em NeurIPS}.

\bibitem[Hagberg et~al., 2008]{hagberg2008exploring}
Hagberg, A., Swart, P., and S~Chult, D. (2008).
\newblock Exploring network structure, dynamics, and function using networkx.
\newblock Technical report, Los Alamos National Lab.(LANL), Los Alamos, NM
  (United States).

\bibitem[Hajiramezanali et~al., 2019]{hajiramezanali2019variational}
Hajiramezanali, E., Hasanzadeh, A., Narayanan, K., Duffield, N., Zhou, M., and
  Qian, X. (2019).
\newblock Variational graph recurrent neural networks.
\newblock In {\em Advances in Neural Information Processing Systems}, pages
  10700--10710.

\bibitem[Jin et~al., 2018]{jin2018junction}
Jin, W., Barzilay, R., and Jaakkola, T. (2018).
\newblock Junction tree variational autoencoder for molecular graph generation.
\newblock In {\em ICML}.

\bibitem[Jin et~al., 2020]{jin2020hierarchical}
Jin, W., Barzilay, R., and Jaakkola, T. (2020).
\newblock Hierarchical generation of molecular graphs using structural motifs.
\newblock In {\em International Conference on Machine Learning}, pages
  4839--4848. PMLR.

\bibitem[Kingma and Ba, 2014]{kingma2014adam}
Kingma, D.~P. and Ba, J. (2014).
\newblock Adam: A method for stochastic optimization.
\newblock {\em arXiv preprint arXiv:1412.6980}.

\bibitem[Kingma and Welling, 2014]{kingma2013auto}
Kingma, D.~P. and Welling, M. (2014).
\newblock Auto-encoding variational bayes.
\newblock In {\em ICLR}.

\bibitem[Li et~al., 2018]{li2018learning}
Li, Y., Vinyals, O., Dyer, C., Pascanu, R., and Battaglia, P. (2018).
\newblock Learning deep generative models of graphs.
\newblock In {\em ICLR}.

\bibitem[Liao et~al., 2019]{liao2019efficient}
Liao, R., Li, Y., Song, Y., Wang, S., Hamilton, W., Duvenaud, D.~K., Urtasun,
  R., and Zemel, R. (2019).
\newblock Efficient graph generation with graph recurrent attention networks.
\newblock In {\em Advances in Neural Information Processing Systems}, pages
  4257--4267.

\bibitem[Liu et~al., 2019]{liu2019graph}
Liu, J., Kumar, A., Ba, J., Kiros, J., and Swersky, K. (2019).
\newblock Graph normalizing flows.
\newblock {\em arXiv preprint arXiv:1905.13177}.

\bibitem[Maniu et~al., 2019]{maniu2019experimental}
Maniu, S., Senellart, P., and Jog, S. (2019).
\newblock An experimental study of the treewidth of real-world graph data.
\newblock In {\em 22nd International Conference on Database Theory (ICDT
  2019)}. Schloss Dagstuhl-Leibniz-Zentrum fuer Informatik.

\bibitem[Murphy, 2012]{murphy2012machine}
Murphy, K.~P. (2012).
\newblock {\em Machine learning: a probabilistic perspective}.
\newblock MIT press.

\bibitem[Niu et~al., 2020]{niu2020permutation}
Niu, C., Song, Y., Song, J., Zhao, S., Grover, A., and Ermon, S. (2020).
\newblock Permutation invariant graph generation via score-based generative
  modeling.
\newblock {\em arXiv preprint arXiv:2003.00638}.

\bibitem[Pennycuff et~al., 2017]{pennycuff2017temporal}
Pennycuff, C., Aguinaga, S., and Weninger, T. (2017).
\newblock A temporal tree decomposition for generating temporal graphs.
\newblock {\em arXiv preprint arXiv:1706.09480}.

\bibitem[Polishchuk, 2020]{polishchuk2020crem}
Polishchuk, P. (2020).
\newblock Crem: chemically reasonable mutations framework for structure
  generation.
\newblock {\em Journal of Cheminformatics}, 12(1):1--18.

\bibitem[Rarey and Dixon, 1998]{rarey1998feature}
Rarey, M. and Dixon, J.~S. (1998).
\newblock Feature trees: a new molecular similarity measure based on tree
  matching.
\newblock {\em Journal of computer-aided molecular design}, 12(5):471--490.

\bibitem[Shi et~al., 2020]{shi2020graphaf}
Shi, C., Xu, M., Zhu, Z., Zhang, W., Zhang, M., and Tang, J. (2020).
\newblock Graphaf: a flow-based autoregressive model for molecular graph
  generation.
\newblock {\em arXiv preprint arXiv:2001.09382}.

\bibitem[Simonovsky and Komodakis, 2018]{simonovsky2018graphvae}
Simonovsky, M. and Komodakis, N. (2018).
\newblock Graphvae: Towards generation of small graphs using variational
  autoencoders.
\newblock In {\em ICANN}.

\bibitem[Su et~al., 2019]{su2019graph}
Su, S.-Y., Hajimirsadeghi, H., and Mori, G. (2019).
\newblock Graph generation with variational recurrent neural network.
\newblock {\em arXiv preprint arXiv:1910.01743}.

\bibitem[Veli{\v{c}}kovi{\'c} et~al., 2017]{velivckovic2017graph}
Veli{\v{c}}kovi{\'c}, P., Cucurull, G., Casanova, A., Romero, A., Lio, P., and
  Bengio, Y. (2017).
\newblock Graph attention networks.
\newblock {\em arXiv preprint arXiv:1710.10903}.

\bibitem[You et~al., 2018a]{you2018graphConv}
You, J., Liu, B., Ying, Z., Pande, V., and Leskovec, J. (2018a).
\newblock Graph convolutional policy network for goal-directed molecular graph
  generation.
\newblock In {\em Advances in neural information processing systems}, pages
  6410--6421.

\bibitem[You et~al., 2018b]{you2018graph}
You, J., Ying, R., Ren, X., Hamilton, W.~L., and Leskovec, J. (2018b).
\newblock Graphrnn: generating realistic graphs with deep auto-regressive
  models.
\newblock In {\em ICML}.

\bibitem[Zang and Wang, 2020]{zang2020moflow}
Zang, C. and Wang, F. (2020).
\newblock Moflow: an invertible flow model for generating molecular graphs.
\newblock In {\em Proceedings of the 26th ACM SIGKDD International Conference
  on Knowledge Discovery \& Data Mining}, pages 617--626.

\end{thebibliography}

\newpage

\appendix
\onecolumn

\section{CANONICAL NAMES FOR TREE NODES} 
\label{sec:can-name}

Algorithm~\ref{alg:canonName} shows the steps to calculate the canonical name of each node in a rooted tree. We use the canonical name of the root as the canonical name of the rooted tree. It can be proven that two rooted trees are isomorphic \emph{iff} they have the same canonical name for their roots \citep{aho1974design,campbell1991tree}. 

\textbf{Canonical Root:} We choose the center of the tree as the canonical root of the tree. In the case that the tree has two centers, we choose the one that creates the greater canonical name for the tree.

Two trees are isomorphic if they have the same canonical name resulting from rooting them from their canonical root~\citep{aho1974design,campbell1991tree}. Figure~\ref{fig:PLR_illustrate}(a) illustrates canonical naming for an example tree rooted from its canonical root.

\begin{algorithm}[H]
\label{alg:canonName}
\SetAlgoLined
\uIf{$v$ is leaf}{return $``ab"$}
\Else{
$childrenNames = EmptyList()$\;
    \For{ each $u$ in $children[v]$}{
        $childrenNames.append(canonicalName(u))$\;
    }
    $sort(childrenNames)$\;
    return $``a"+childrenNames.toString()+``b"$\;
}

\caption{$getCanonicalName(v)$}
\end{algorithm}

We use the following characteristics of the canonical names in our proofs:
\begin{enumerate}
    \item If we define the subtree of the descendant nodes of a node, e.g. $v$, as $T_v$, and denote the canonical name of $v$ as $canonicalName(v)$, then we have $|canonicalName(v)| = 2|T_v|$. It means that the length of the canonical name of $v$ is two times the number of nodes in the tree of the descendants of $v$.
\item Each canonical name has an equal number of "a"s and "b"s. Also, in each absolute prefix of a canonical name (prefix shorter than the canonical name), the number of "a"s is greater than the number of "b"s. (In other words if we change "a"s to "("s and "b"s to ")" we have a valid parenthesis sequence.) 
\item The canonical names of the children of a node can be uniquely determined from the canonical name of a node.
\end{enumerate}

\section{TREE DECOMPOSITION ALGORITHM}
\label{appendix:td-alg}
Finding the optimal tree decomposition is well-known to be an NP-hard problem. However, there are several algorithms designed for finding tree decompositions that work well both in terms of width and time complexity \citep{bodlaender2010treewidth}. 
JT-VAE uses an algorithm rooted in chemistry~\citep{rarey1998feature} 
which relies on the fact that the number of unique possible structures for the clusters can be reduced to a small value ($780$ in JT-VAE). However, this is a domain-specific algorithm and can not be extended to the general graphs. Further, in most benchmark graph datasets, such as community and community-small, any subgraph is likely to be generated.
Thus, reduction in these cases is not feasible. Knowing this problem, we use the Fill-in algorithm introduced in \citep{bodlaender2010treewidth} with the standard implementation of the NetworkX library~\citep{hagberg2008exploring} for this decomposition. However, we make some modifications in the algorithm to make our general algorithm work better both in terms of complexity of the generation and complexity of learning. To this end, we define a minimal tree decomposition as:

\textbf{Minimal Tree Decomposition:} We call a tree decomposition $\mathcal{T}$ minimal \textit{iff} there are no two adjacent supernodes such as $i$ and $j$ that we have $X_i \subseteq X_j$ or $X_j \subseteq X_i$.

To obtain the minimal tree decomposition, after using the fill-in algorithm, we traverse the tree decomposition again by the order of the canonical names of the nodes, and we merge each two neighbor supernodes if any of them is the subset of the other. For merging, we make a new supernode with the union of the two supernodes, and we contract the edge between these two nodes. We continue the merging process until we end up with a minimal tree decomposition. We do not use other characteristics of the fill-in algorithm. So, this algorithm can be replaced by any other algorithm used for different applications.

\section{MODEL DETAILS}
\label{appendix:model_details}

\subsection{PLR}
\label{appendix:PLR-boundaries}

In this section first we explain the algorithm to regenerate a tree from a $\mathsf{PLR}$. Then we will explain how we can find the boundaries for the next length in a $\mathsf{PLR}$ given a prefix of it.

The $\mathsf{PLR}\_\mathsf{to}\_\mathsf{Tree}$ algorithm starts with a node, which is the root of the tree. Then, we iteratively add paths and find the next node to extend. Pseudocode for this algorithm is in Alg.~\ref{alg:PLR_to_Tree}.

\begin{algorithm}[H]
\label{alg:PLR_to_Tree}
\SetAlgoLined
$root := 0$\;
$T := (root)$\;
$extending\_node := root$\;
$next\_ID := 1$

\For{ $i$ from $1$ to $r-1$}{
    In $T$ add a path of length $\ell_i$ from $extending\_node$ with node IDs: $next\_ID$ to $next\_ID + \ell_i - 1$ \;
    $next\_ID = next\_ID + \ell_i$\;
    \uIf{$\ell_i > 0$}{$extending\_node = next\_ID - 2$}
    \Else{$extending\_node = parent_T(extending\_node)$}
}
return $T$, $extending\_node$\;

 \caption{$\mathsf{PLR}\_\mathsf{to}\_\mathsf{Tree}((\ell_1, \ldots, \ell_r))$}
\end{algorithm}

In the generated rooted tree, we define the left brother of node $v$ as the child of its parent which is generated immediately before it. If there is no such a node we call $v$ the first child of its parent. In other words, we sort the children of each node with their IDs, and the left brother of each node comes in its parent's list immediately before it. Now, a $\mathsf{PLR}$ is valid if it satisfies these conditions:
\begin{enumerate}
    \item Based on the definition, $\ell_i \geq 0$ for all $i$.
    All nodes in a $\mathsf{PLR}$ have a parent except the root, and the $\ell_r=0$  indicates the end of the $\mathsf{PLR}$.
    Hence, an absolute prefix of a $\mathsf{PLR}$  itself is not a complete $\mathsf{PLR}$.
    \item The returned $extending\_node$ in Algorithm~\ref{alg:PLR_to_Tree} should be the root and $\ell_r$ should be zero to assert that the $\mathsf{PLR}$ reaches the end.
    \item The root should be the canonical root of $T$.
    \item For each node its canonical name should be smaller than or equal to its left brother's name (if exists). This condition ensures  that in the process of creating $\mathsf{PLR}(T)$,  nodes are visited with respect to their $node\_ID$ order. The immediate result of this is that $\mathsf{PLR}(T) := \mathsf{DFS}\_\mathsf{PLR}(root,0) = (\ell_1, \ldots, \ell_r)$.
\end{enumerate}

Given a prefix of the $\mathsf{PLR}$, e.g., $(\ell_1, \ldots, \ell_{i-1})$ using the $\mathsf{PLR}\_\mathsf{to}\_\mathsf{Tree}$ algorithm, we can find the tree generated until now, \emph{i.e.}, $T$, and the node that the next path will be extended from, \emph{i.e.}, $extending\_node$. For boundaries of the next path we shall not violate the aforementioned conditions:

\begin{enumerate}
    \item For the first and the second conditions: We should have $\ell_i \geq 0$. Also, if $extending\_node = root$, then $\ell_i$ can be zero \textit{iff} generation finishes at this step.
    \item For the third condition: If $\ell_i$ is the second path extended from the root, we should have $\ell_1 -1 \leq \ell_i \leq \ell_1$. This ensures that the root is the center of the tree. If we select that $\ell_i = \ell_1$, the tree has just one center, otherwise it will have two centers. In the second case, to make sure that the correct center is the root, we need to calculate two canonical names: 1) the canonical name of the first node of the root, 2) the canonical name of the root if we remove its first child and its subtree. Now name 1 should be greater than name 2. Adding new nodes in a subtree can only make its ancestor's name greater. So, if this state happens for the next lengths in $\mathsf{PLR}$ or $\ell_{j>i}$, it gives an upper bound on the path that the name 2 can not exceed name 1.
    \item For the fourth condition: Similar to the previous one, adding new nodes can just increase the canonical name of the ancestors. In this case, we should check $extending\_node$  and its ancestors for the maximum length we can add such that after adding a new node none of the nodes gets greater canonical name than its left brother.
\end{enumerate}

The aforementioned conditions gives the boundaries for $\ell_i$. If $\ell_i$ is not the second path extending from the $root$, the lower bound is always zero.

\subsection{Optimizing}
We used the Adam optimizer in all of the training processes \citep{kingma2014adam}. We started with initial learning rate $0.0005$ in the tree generator model, and $0.001$ in the node sharing and node adding models. For all of the models, the learning rate is reduced by a factor of $0.5$ if in $25$ consecutive epochs the validation loss does not decrease.

\subsection{Early Stopping}
For standard training in all of the models, we stop training if the loss on the validation data does not decrease in $50$ consecutive epochs. Based on our experimental studies, for most of the datasets, the training process ends before the $500$th epoch. Finally, we use the saved model with the best validation loss.

For overfitting, we track the loss on the training data. If the loss on the training data does not decrease for $50$ consecutive epochs, we stop the training. Finally, we use the saved model with the best training loss.

\subsection{Hyperparameters}
For training the tree generator, we set the hidden size of the tree encoder to $32$ for the small graph datasets and the lobster trees. For the normal sized graph datasets, we set this value to $64$. In the tree encoder model, we always set the size of the hidden layers of the MLP equal to the hidden size of the tree encoder.

Here, we explain the settings for the node sharing and node adding models. For the small datasets, we use two layers of graph attention network for encoding the graph, and we set the encoding size of these networks to $32$. We also use tree encoding of size $16$ in these models. For the normal sized datasets, we use four layers of graph attention network with encoding size of $64$ and tree encoding size of $32$.

\subsection{Resources and Training Time}
All of the experiments are conducted using a single Nvidia GTX 1080 Ti GPU. Total time required for the training for each of datasets are as: 1) Ego-small: ~2 hours; 2) Community-small: ~3 hours; 3) Ego: ~6 hours; 4) Community: ~12 hours. For the tree generation for Lobster trees the total time is less than an hour for the standard training.

\section{PROOFS}
\label{appendix:proofs}

In this section, we will prove the theorems and propositions that came in the paper.

\textbf{Lemma 6.} Each two isomorphic trees have the same $\mathsf{PLR}$.

\textbf{Proof.}
If we have two trees $T_1$ and $T_2$, if they are isomorphic according to \citep{aho1974design}, we have the same name for them if we make each of them from its canonical root. According to the characteristics of the canonical names, both trees have the same number of nodes as the roots have the same canonical name. We define a bijective mapping, e.g., $f$, between the first and second tree's nodes, inductively. If trees have just one node, there is an obvious mapping between their roots. For more than one node, the roots have at least one child. As the roots have the same canonical name, we can uniquely find the canonical names of their children. As a result, they have the same multi-set of names for their children. If root nodes are $v_1$ and $v_2$ we set $f(v_1) = v_2$; and then sort the children of each of them by decreasing order of their canonical name. Then child $i$ of $v_1$ has the same canonical name with the child $i$ of $v_2$. Thus, according to the induction, we can make a bijective mapping between the subtree of nodes of child $i$ of $v_1$ and $v_2$. These mappings make a bijective mapping between the nodes of the two trees.

Now, when we are doing $\mathsf{DFS}\_\mathsf{PLR}$ on $T_1$ and $T_2$; at step number $j$ of both algorithms if we are at node $u_j$ on $T_1$, then we are on node $f(u_j)$ of $T_2$ as they have the same structure and the ordering of the nodes in the $\mathsf{DFS}\_\mathsf{PLR}$ is the same. The value of the $l$ is the same on each corresponding step of the algorithm on two trees. The latter claim can also be proved using induction. At step 1, both algorithms have $l=0$, if they have the same value till step $i$, for the next step, if one node calling its child, it would add $l$ value by one and call its child that we know that is a bijection of the similar node from the other tree, so they would have the same value of $l$. If the next step is returning from one finished (black) node, the next step would be with $l=0$ for both of the trees, and so when we are adding path lengths to the $\mathsf{PLR}$, we would have the same values. And as a result, we would have the same $\mathsf{PLR}$ for $T_1$ and $T_2$.
$\blacksquare$

\textbf{Lemma 7.} The size of the $\mathsf{PLR}$ for a tree with $r$ nodes is $r$.

\textbf{Proof.} According to the Algorithm \ref{alg_PLR}, as the tree is a connected graph, the DFS algorithm would traverse all the nodes of the tree. In the DFS procedure, we add exactly one length to the $\mathsf{PLR}$ for each node, so the length of the $\mathsf{PLR}$ is equal to the number of the nodes of the tree. Furthermore, for each leaf of the tree, we have one and exactly one non-zero item in the $\mathsf{PLR}$, and for each non-leaf node of the tree, we have one and exactly one zero in the $\mathsf{PLR}$.$\blacksquare$

\textbf{Lemma 8.} Given a $\mathsf{PLR}$ we can uniquely find the tree the $\mathsf{PLR}$ was generated from.

\textbf{Proof.} We have previously explained how to create tree from the $\mathsf{PLR}$ in $\mathsf{PLR}\_\mathsf{to}\_\mathsf{Tree}$.
Now, if we consider the steps of reconstructing from $\mathsf{PLR}$ and we assume that we are adding the path one node at a time and move toward a leaf and add node till we finish it. This process is the same as the order $\mathsf{DFS}\_\mathsf{PLR}$ passes on the final created tree. At each step, we are moving from a node to its child with the greatest canonical name (which we make sure a valid $\mathsf{PLR}$ has this property by setting valid boundaries at each step). And when we reach a leaf, a path is finished, and when we finish a non-leaf node and want to move to the parent, zero is seen in the $\mathsf{PLR}$. As a result, if we find the $\mathsf{PLR}$ using $\mathsf{DFS}\_\mathsf{PLR}$ for the recently constructed tree with the canonical root and the canonical ordering over the nodes, we would end up by the initial $\mathsf{PLR}$. As a result, the defined process is the exact reverse of the $\mathsf{DFS}\_\mathsf{PLR}$ process. There is a bijection between the steps of $\mathsf{DFS}\_\mathsf{PLR}$ and the $\mathsf{PLR}\_\mathsf{to}\_\mathsf{Tree}$, so the initial tree is unique. Suppose there was another tree that generated the same $\mathsf{PLR}$, according to the algorithm, we could have reached that tree using the reversing algorithm, which we know is impossible according to unique actions in the reversing process. Thus, the initial tree is unique. $\blacksquare$

\textbf{Theorem 1.} There is a bijection between the $\mathsf{PLR}$s of length $r$ 
and the space of the isomorphic classes of trees with $r$ nodes.

\textbf{Proof.} According to Lemma 6, every two trees in the same isomorphism class have the same $\mathsf{PLR}$. And according to Lemma 8, each $\mathsf{PLR}$ can be reversed to find the initial rooted tree with the canonical root and canonical ordering over its nodes, which can be considered as the unique representative of that class of isomorphic trees. According to Lemma 7. the length of the $\mathsf{PLR}$ for a tree with $r$ nodes is $r$.As a result, there is a bijection between the $\mathsf{PLR}$s of length $r$ and the space of the isomorphic classes of trees with $r$ nodes. $\blacksquare$

\textbf{Theorem 2.} Each valid $\mathsf{PLR}$ can be generated by the proposed model, and every $\mathsf{PLR}$ generated using this model is valid.

\textbf{Proof.} Using $\mathsf{PLR}$ boundaries defined in section~\ref{appendix:PLR-boundaries}, we know that given $(\ell_1, \ldots, \ell_{i-1})$, what is the boundaries for the $\ell_i$, let us assume that given these conditions we have $a \leq \ell_i \leq b$. First, we prove that given any of these values there is a valid $\mathsf{PLR}$ that $(\ell_1, \ldots, \ell_{i-1}, \ell_i)$ is a prefix of it. We can prove this by constructing a valid  $\mathsf{PLR}$ by extending this prefix. For constructing, if the second child of the root is already generated, we just add $0$ until we reach back to the root, and finally, by adding another $0$, we end up with a valid $\mathsf{PLR}$. If the second child is not generated yet, we can add $0$ until we reach back to the root, then we add a path of length $\ell_1$, and again we continue to add $0$ until we reach back to the root, and by a final $0$ we will have a valid $\mathsf{PLR}$. As a result, for any value in the boundary, the sequence is a prefix of at least one valid $\mathsf{PLR}$, so every $\mathsf{PLR}$ generated by the model should be valid.

Now, we prove that the model can generate every valid $\mathsf{PLR}$. Using the Softmax function ensures that any value inside the boundary has a positive probability to be generated. So, having a positive probability of being generated is equal to having the correct value of the next path length inside the boundaries we have introduced. We prove this section by contradiction: let us assume that there is a $\mathsf{PLR} = (\ell_1, \ldots, \ell_r)$ such that the model can not produce this $\mathsf{PLR}$. If the model can not generate this sequence there is an index like $i \leq r$ such that $(\ell_1, \ldots, \ell_{i-1})$ can be produced by the model, but $\ell_i$ is not in the boundary limits that algorithm finds using $(\ell_1, \ldots, \ell_{i-1})$. If the boundaries found for this path are $a$ and $b$, $a$ can be greater than zero only if the $extending\_node$ at this step is the root. So, if $\ell_i < a$, the condition that the root is the canonical root is violated, so the given $\mathsf{PLR}$ is not a valid $\mathsf{PLR}$. In other cases we always have $a=0$ so $\ell_i \geq a$. The only other state that $\ell_i$ can not be within this boundary is $\ell_i > b$. In this case, according to how we find $b$, there should be a node from $extending\_node$ or its ancestors such that by adding this path, the canonical name for that node becomes greater than its left brother's canonical name. Or the condition that the root is the canonical root (right center) will be violated. According to the DFS order, in both conditions generation of the subtree of the left brother, or the first child of the root is already finished and their canonical name will not change by adding new paths. And the canonical name of the ancestor node or the root will always be increasing by adding a new path, so the conditon that canonical name of each node should be less or equal to its left brother, or the condition that canonical name of the first child should be greater or equal to the canonical name came from removing of the first child and its subtree will not satisfy by extending this prefix. So, the given sequence is not a valid $\mathsf{PLR}$, so this is in contradiction with the initial assumption, so the model can generate every valid $\mathsf{PLR}$. $\blacksquare$

\textbf{Proposition 3.} Given a graph $G$ with $n$ nodes and a minimal tree decomposition of it, e.g., $\mathcal{T} = (T,C)$, with width $k$, then $r \leq n-k+1$, where $r$ is the number of nodes in $T$.

\textbf{Proof.}  
Let $i_1 = \arg \max_i |X_i|$, so $|X_{i_1}| = k$. Starting with $\{i_1\}$ we add other super nodes of $\mathcal{T} = (\mathcal{X} , T)$ step by step to this set. At step $j$ we select one new super node ${i_j}$, which is a neighbor of one of the supernodes $i_1, i_2, \cdots, i_{j-1}$. With this order of selecting supernodes, induced subgraph of $T$ on these supernodes is a connected subtree of $T$. From another perspective, this process is like making ${i_1}$ as the root of the tree and starting adding other nodes in a way that we always have a connected subtree of $T$. We define $parent(i)$ as parent of the supernode $i$ in the rooted tree with root ${i_1}$. Let $G_j$ be the induced subgraph of $G$ with nodes $V_j = X_{i_1} \cup X_{i_2} \cup \cdots \cup X_{i_j}$ and let $V_0 = \emptyset$. As $T$ is minimal $X_{i_j} \not\subset X_{parent(i_j)}$, and $X_{i_j} \cap V_{j-1} \subseteq X_{parent(i_j)}$, so $X_{i_j} - V_{j-1} \neq \emptyset$, so $V_j \neq V_{j-1}$ and obviously $V_{j-1} \subset V_j$. As a result we have $|V_j| > |V_{j-1}|$. Now we know that $|V_1| = k$ and $|V_r| = n$ as $G_r = G$, and also for each $j$, $|V_j| > |V_{j-1}|$ , so we should have $r \leq n-k+1$. $\blacksquare$

\textbf{Theorem 4.} Generating a graph with $n$ nodes using a tree decomposition with width $k$ 
has worst-case complexity of $O(nk)$ with respect to the decision steps.

\textbf{Proof.} 
We can see from proposition three that the total number of steps for generating a tree is linear regarding the number of nodes of the initial graph. For the node sharing steps, we have $O(n)$ supernodes, and for each supernode (e.g., supernode $i$), we decide which nodes from its parent are shared with that node. So, we conduct $|X_{parent(i)}|$ decisions. And from the width of the tree decomposition we know that $|X_{parent(i)}| = O(k)$. So, the total number of decisions for the node sharing is from $O(nk)$. The total number of decisions for adding a node or not is $O(n)$. This is because we have $O(n)$ nodes to add and $O(n)$ stop adding nodes decisions for all of the subgraphs.

For the edge adding decisions, let us assume similar process of adding supernodes from the proof of Propositon 3, but instead of selecting the largest bag for starting we can start with any arbitrary supernode. We also use the same definition on $V_j$ and $parent(i_j)$. Also, we define: $U_j = X_{i_j}-V_{j-1}$. In the step $j$ for generating $C_{i_j}$ we just need to decide about the connections of each node in $U_j$ with other nodes inside the $X_{i_j}$, for each two nodes $u,v \in X_{i_j}-U_j$ we have $u,v \in X_{parent(i_j)}$. As we have generated $C_{parent(i_j)}$ before, so we have already decided about the connection of $u$ and $v$. For each node in $U_j$ as size of $X_{i_j}$ is at most $k$ we need to decide about at most $k-1$ connections. By definition of $U_j$s each node of $G$ is exactly in one of $U_j$s, because $U_1\cup U_2 \cup \cdots \cup U_r = X_{i_1} \cup X_{i_2} \cup \ldots \cup X_{i_r} = \mathcal{V}$ and for each $t$ and $s$, $U_t \cap U_s = \emptyset$, so $|U_1| + |U_2| + \ldots + |U_r| = n$. Therefore, we need to decide about at most $n(k-1) = O(nk)$ edges. So the total number of steps is from $O(nk)$ $\blacksquare$

\textbf{Proposition 5.} For an arbitrary ordering of the nodes $(v_1, \cdots, v_n)$ on a connected graph $G$,
$$tw(G) \leq 2*max_{d \in 1..\operatorname{diam}(G)}
\Big| \{v_{i} | \operatorname{dist}(v_{i}, v_{1})=d\}\Big|.$$

\textbf{Proof.} Let us assume on an arbitrary permutation of the nodes we make a BFS tree rooted from $v_1$, and assume $L_d = \{v_i|\operatorname{dist}(v_i, v_1)=d\}$. We make tree decomposition with supernodes $X_1,X_2,\cdots,X_{\operatorname{diam}(G)}$ such that $X_i = L_{i-1}\cup L_i$, and tree structure $T$ such that $T$ is a path starting from $X_1$, ending with $X_{\operatorname{diam}(G)}$. We can see this decomposition is a tree decomposition, because path is a type of tree, and three conditions of tree decomposition are holding here:
\begin{itemize}
    \item As each node has maximum distance ${\operatorname{diam}(G)}$ from $v_1$, each node should be in exactly one of the $L_i$s. so it should be in at least one of the $X_i$s. As a result we have: $X_1 \cup X_2 \cup \cdots \cup X_{\operatorname{diam}(G)} = \mathcal{V}$.
    \item As we make the $L_i$s using the BFS tree, we know that there are no edges between $L_i, L_j$, which $|i-j|>1$. Therefore, two ends of each edge in $G$ should come together in at least one of the bags.
    \item Each node of $G$ can come in at most two bags of nodes as $\forall i \neq j L_i \cap L_j = \emptyset$, and these two bags should be consecutive, so they are neighbors in the tree decomposition. Thus, the third condition of tree decomposition also holds for this decomposition.
\end{itemize}

As a result, this is a valid tree decomposition, so $tw(G)\leq max_{d=1}^{\operatorname{diam}(G)}|X_d| \leq 2*max_{d=1}^{\operatorname{diam}(G)}|L_d|$

This tree decomposition also shows that limiting the number of edge predictions using BFS ordering can be considered a special case of tree decomposition. Hence, any tree decomposition with better width can outperform BFS regarding the number of edge decisions, which should be determined.$\blacksquare$

\section{STATISTICS FROM DATA}
\label{appendix:stats}
In this section, we perform a statistical analysis of some measures of complexity in our proposed algorithm. Figures \ref{stats:tree-size}, \ref{stats:width}, \ref{stats:cluster-sizes}, and \ref{stats:new-nodes} demonstrate the distribution of different measures of size in tree decomposition and subgraph generation for the following four datasets: Community-small, Ego-small, Community, and Ego. The distributions are calculated by sampling one random permutation per graph for all the graphs in the datasets. 

\subsection{Number of the Supernodes in the Tree Decompositions}
\label{appendix:supernodes-dist}
\begin{figure}[ht!]
\centering
\subfloat[]{\includegraphics[width = 1.4in]{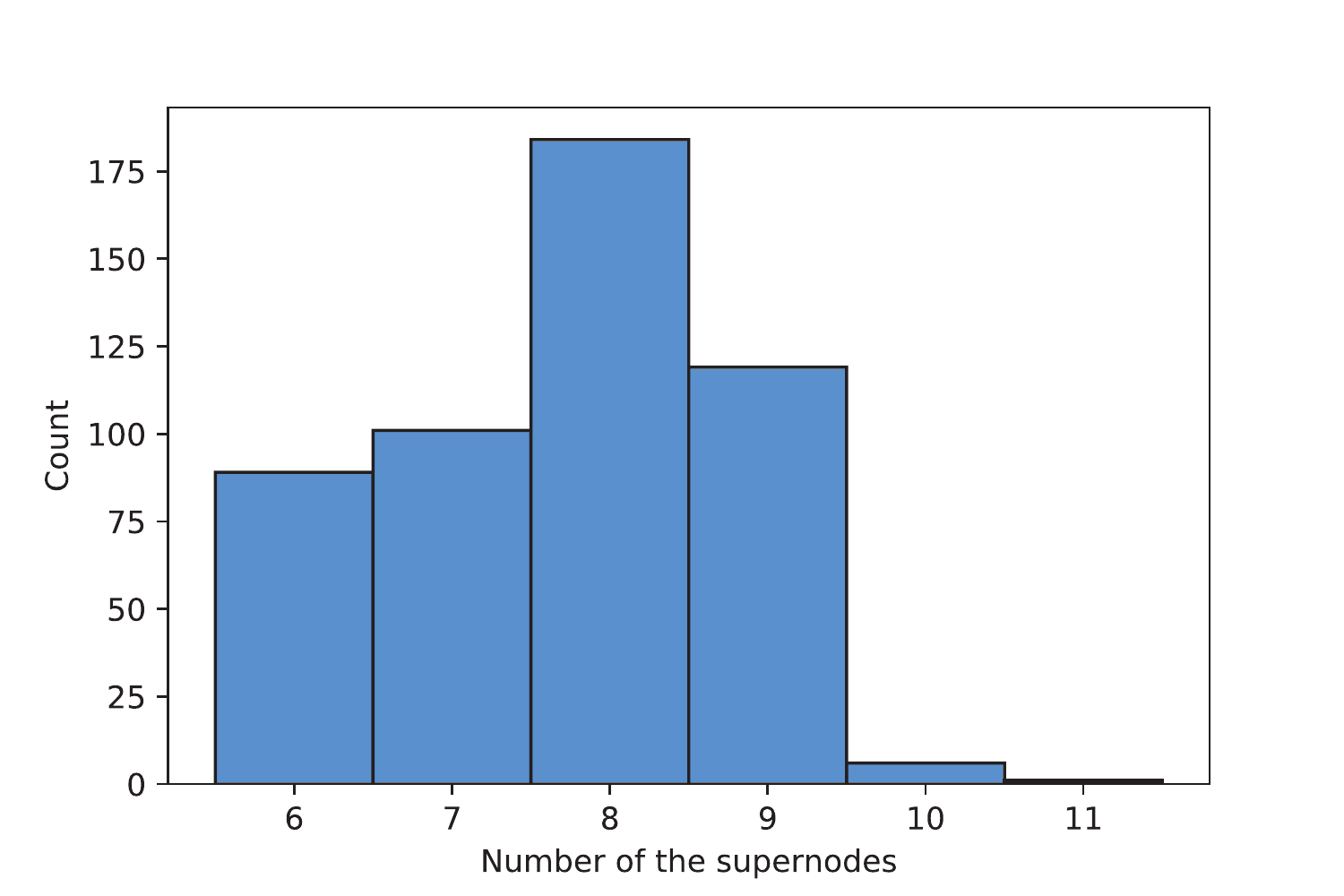}} 
\hspace{-0.1in}
\subfloat[]{\includegraphics[width = 1.4in]{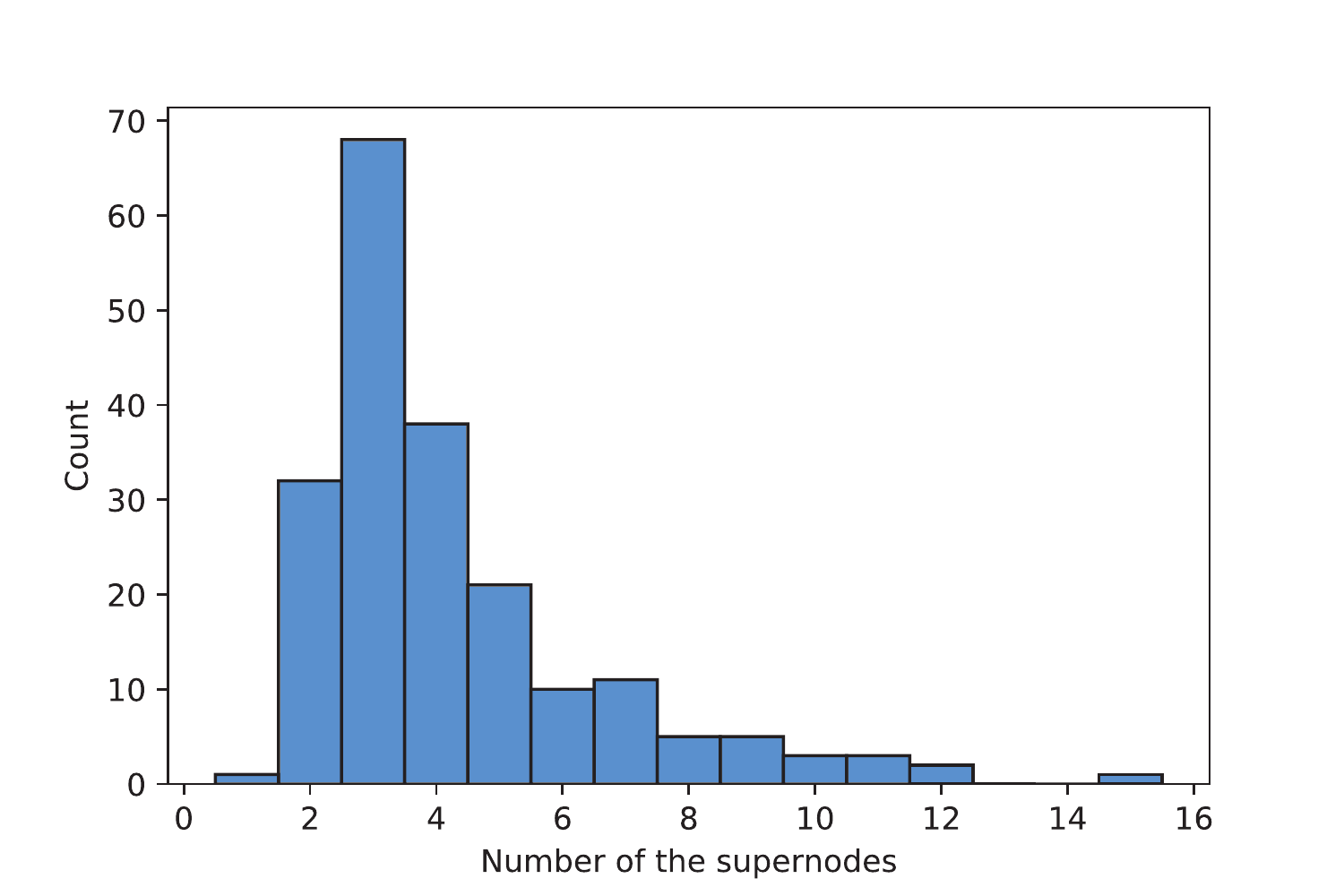}}
\hspace{-0.1in}
\subfloat[]{\includegraphics[width = 1.4in]{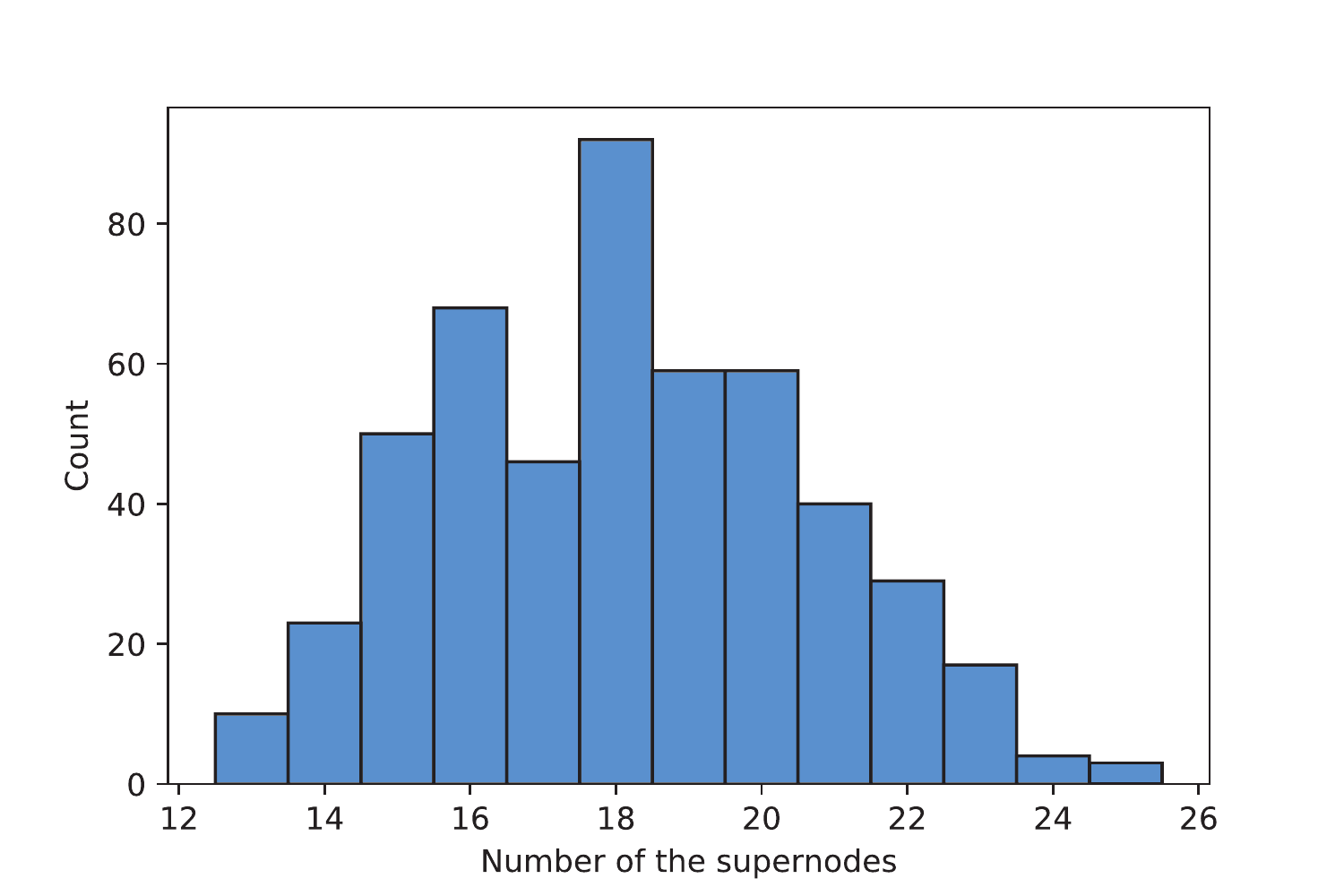}}
\hspace{-0.1in}
\subfloat[]{\includegraphics[width = 1.4in]{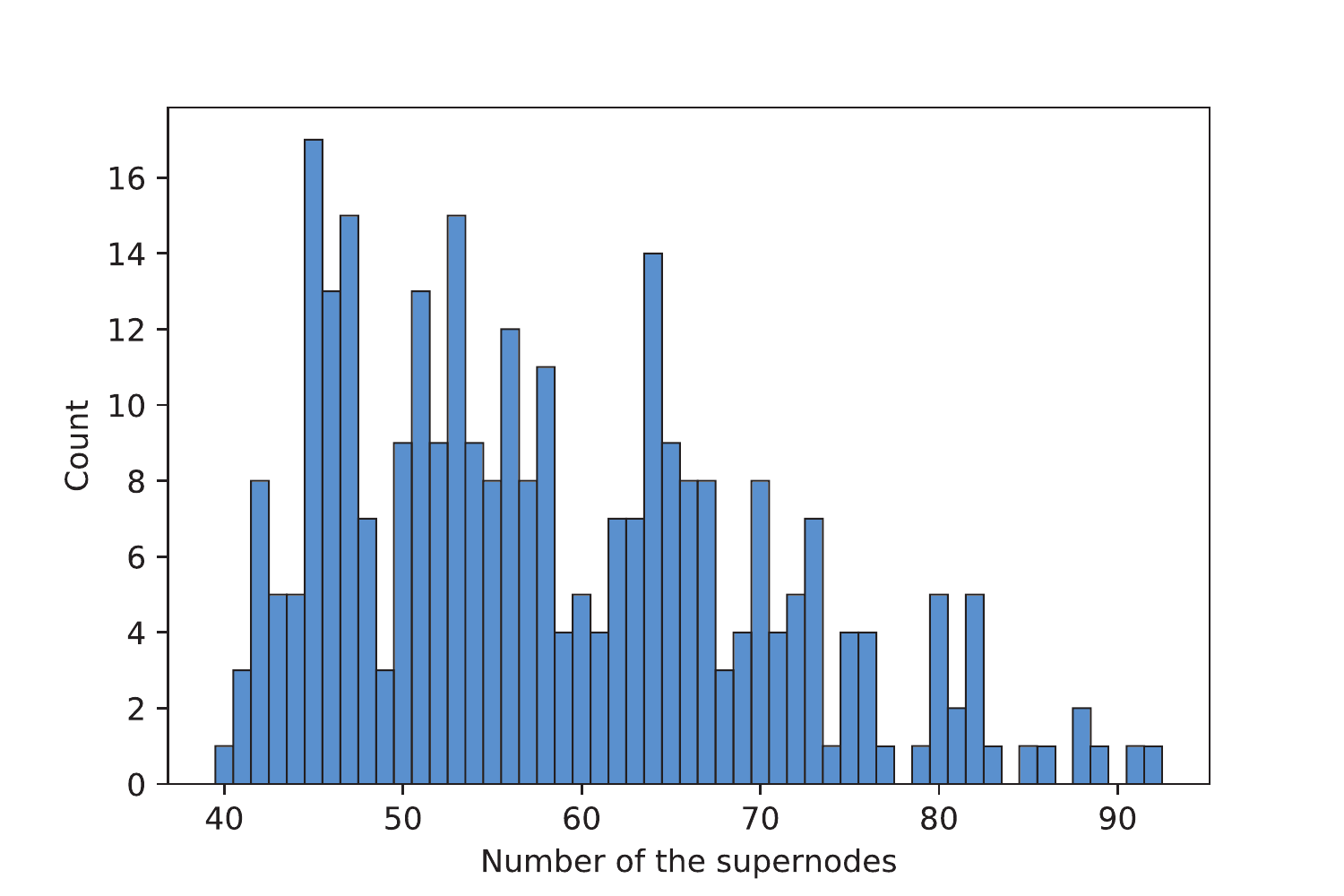}} 
\caption{The distribution of the number of the tree supernodes after tree decomposition. }
\label{stats:tree-size}
\vspace{-0.1in}
\end{figure}

In Figure~\ref{stats:tree-size}, the distribution on the number of the supernodes of the tree after tree decomposition is shown. It can be observed that the number of the supernodes is statistically smaller than the number of nodes of the graphs, which confirms Proposition 3.

\subsection{Width of the Tree Decompositions}
\label{appendix:width-dist}
\begin{figure}[ht!]
\centering
\subfloat[]{\includegraphics[width = 1.4in]{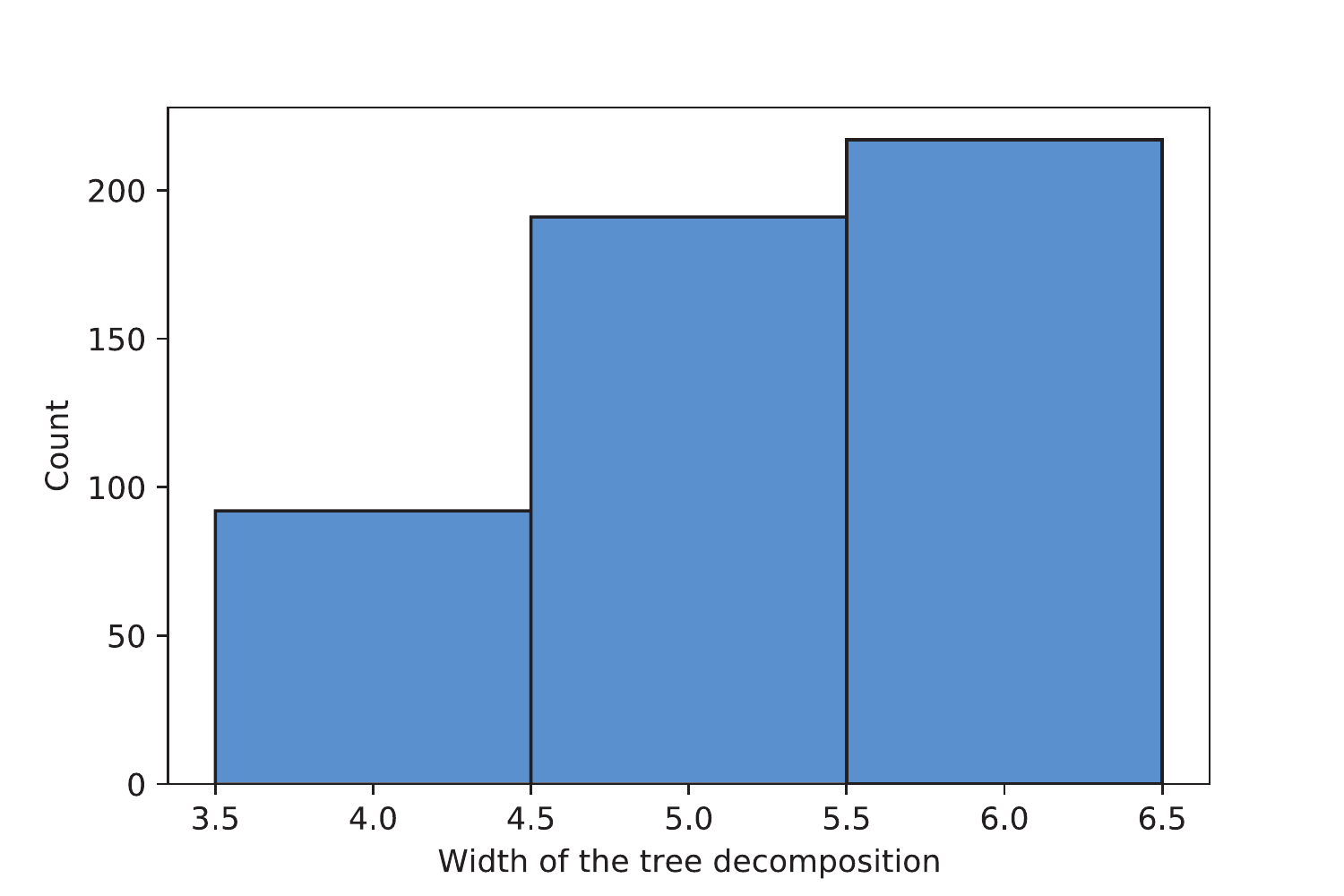}} 
\hspace{-0.1in}
\subfloat[]{\includegraphics[width = 1.4in]{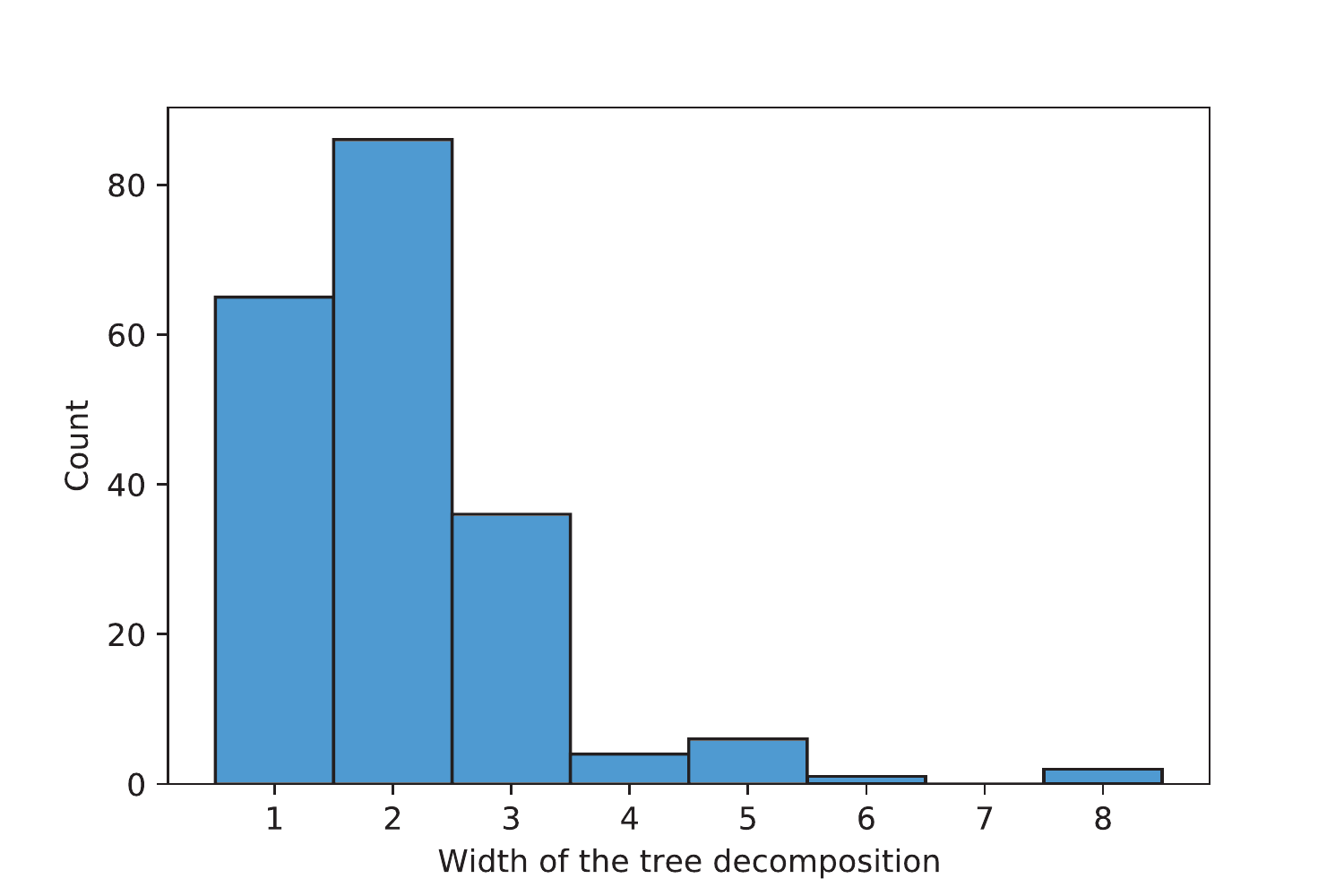}}
\hspace{-0.1in}
\subfloat[]{\includegraphics[width = 1.4in]{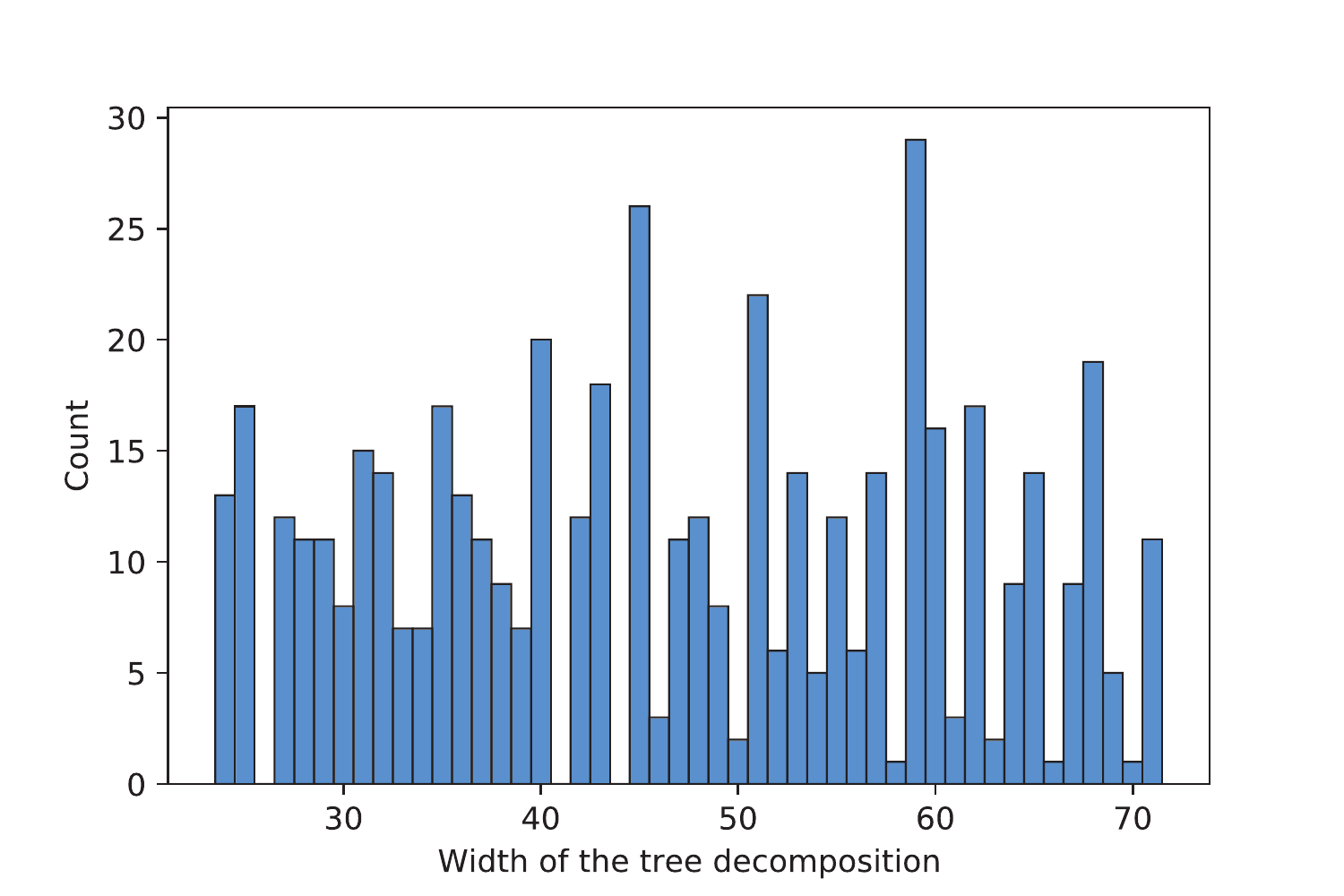}}
\hspace{-0.1in}
\subfloat[]{\includegraphics[width = 1.4in]{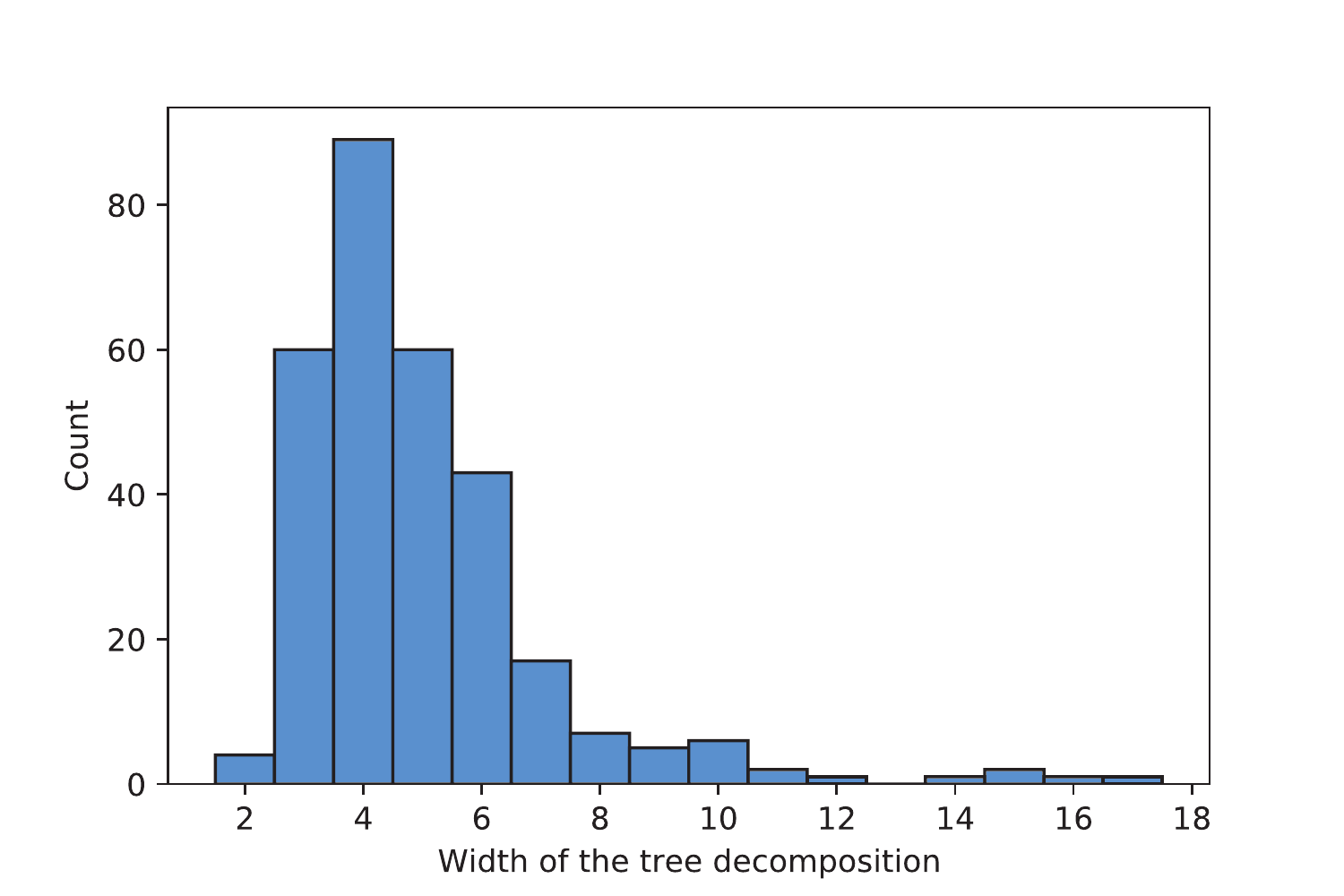}} 
\caption{The distribution of the width of the tree decompositions in the graphs. }
\label{stats:width}
\vspace{-0.1in}
\end{figure}

Figure~\ref{stats:width} shows the distribution of the width of the tree decompositions for the graphs in each dataset. It can be observed that this number is significantly smaller than the number of the graph nodes in most cases.

\subsection{Number of the Nodes in the Subgraphs}
\label{appendix:cluster-size-dist}
\begin{figure}[ht!]
\centering
\subfloat[]{\includegraphics[width = 1.4in]{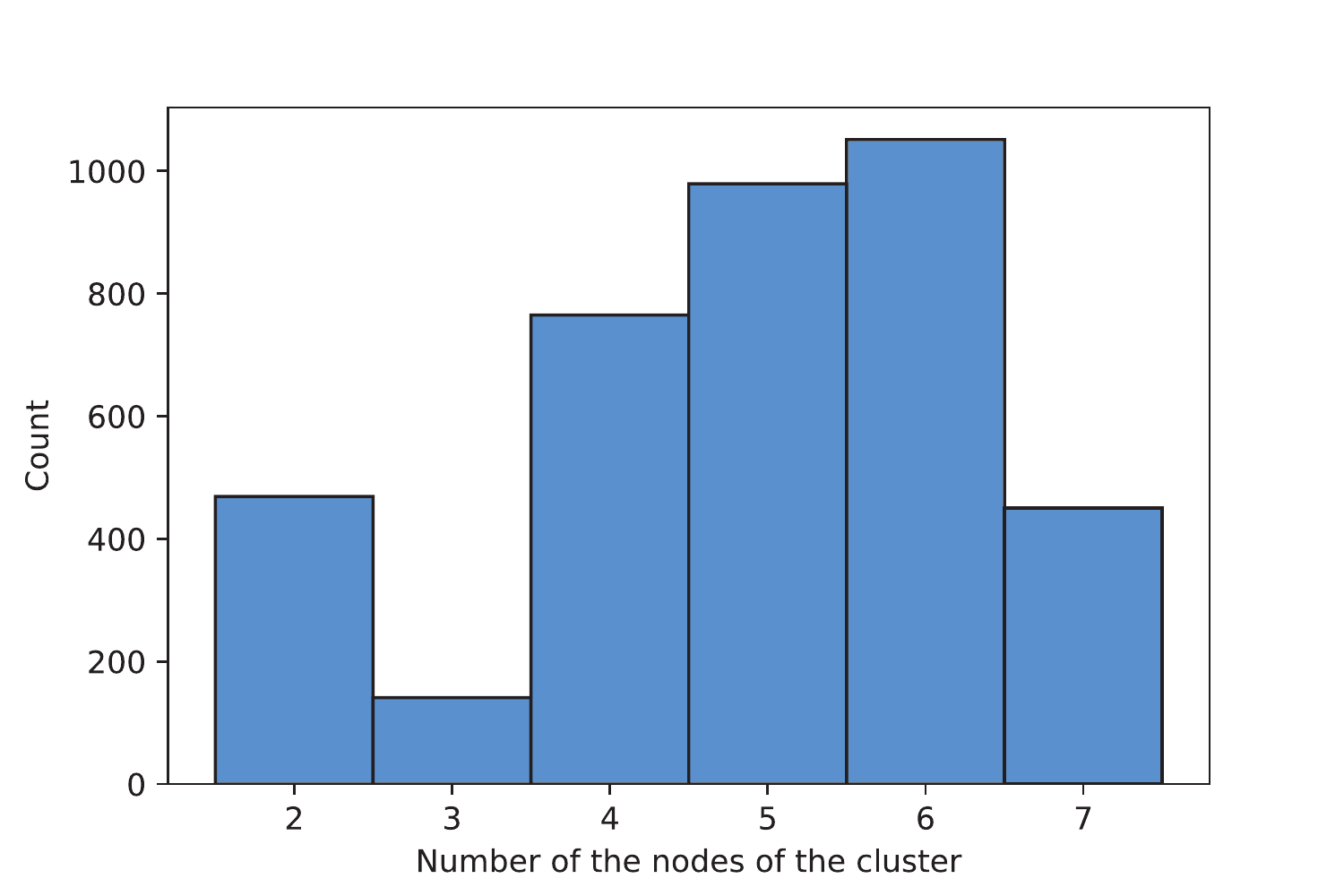}} 
\hspace{-0.1in}
\subfloat[]{\includegraphics[width = 1.4in]{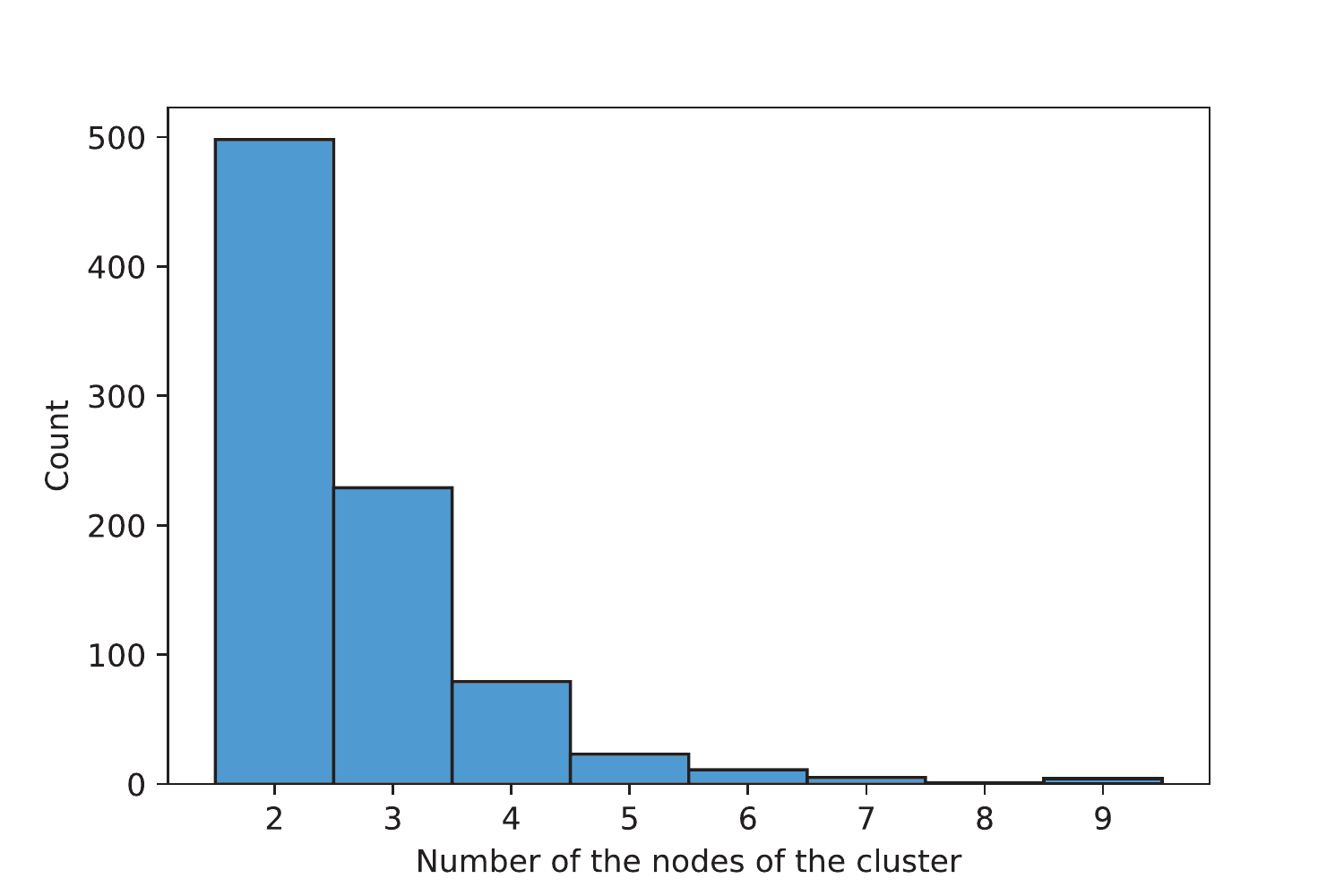}}
\hspace{-0.1in}
\subfloat[]{\includegraphics[width = 1.4in]{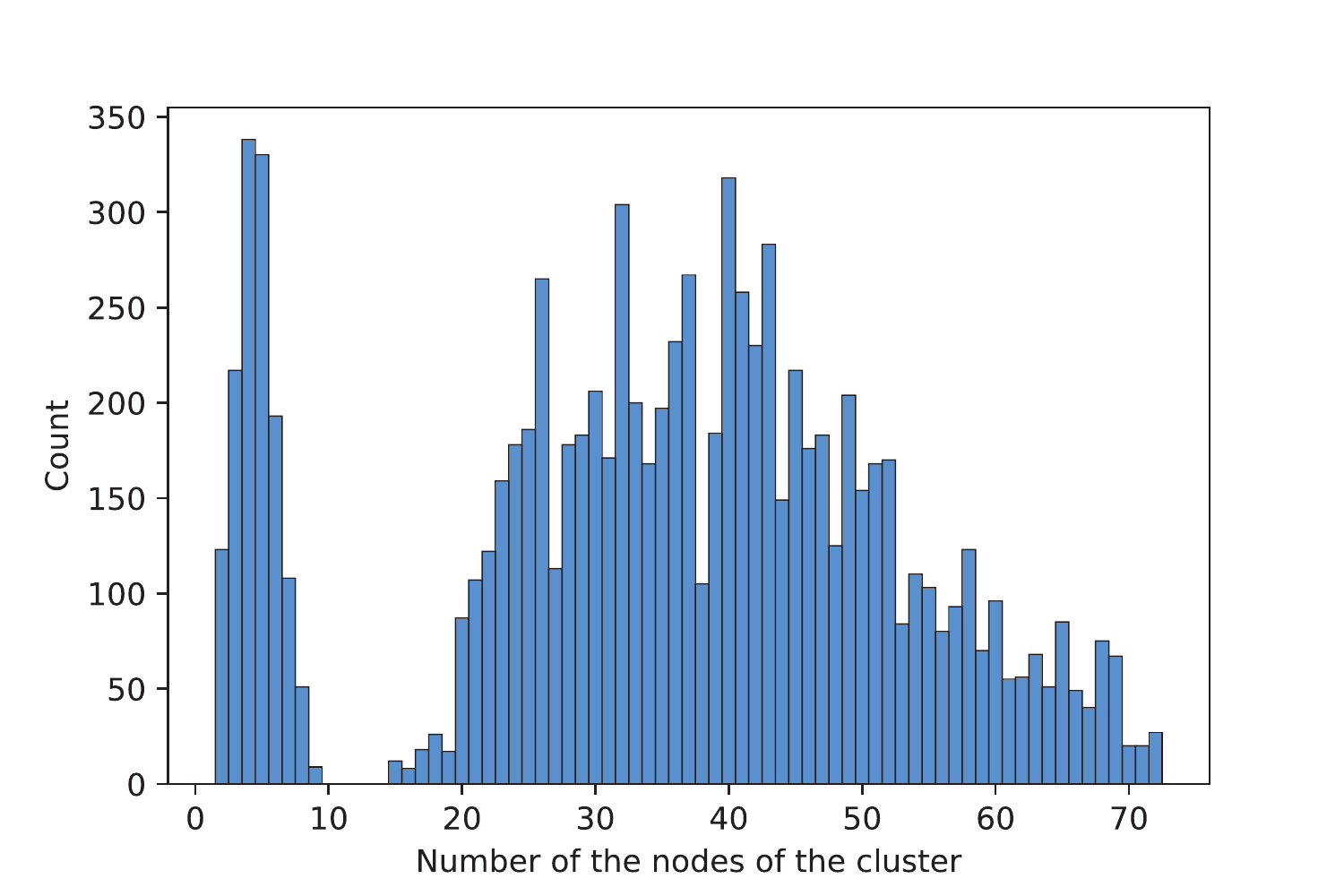}}
\hspace{-0.1in}
\subfloat[]{\includegraphics[width = 1.4in]{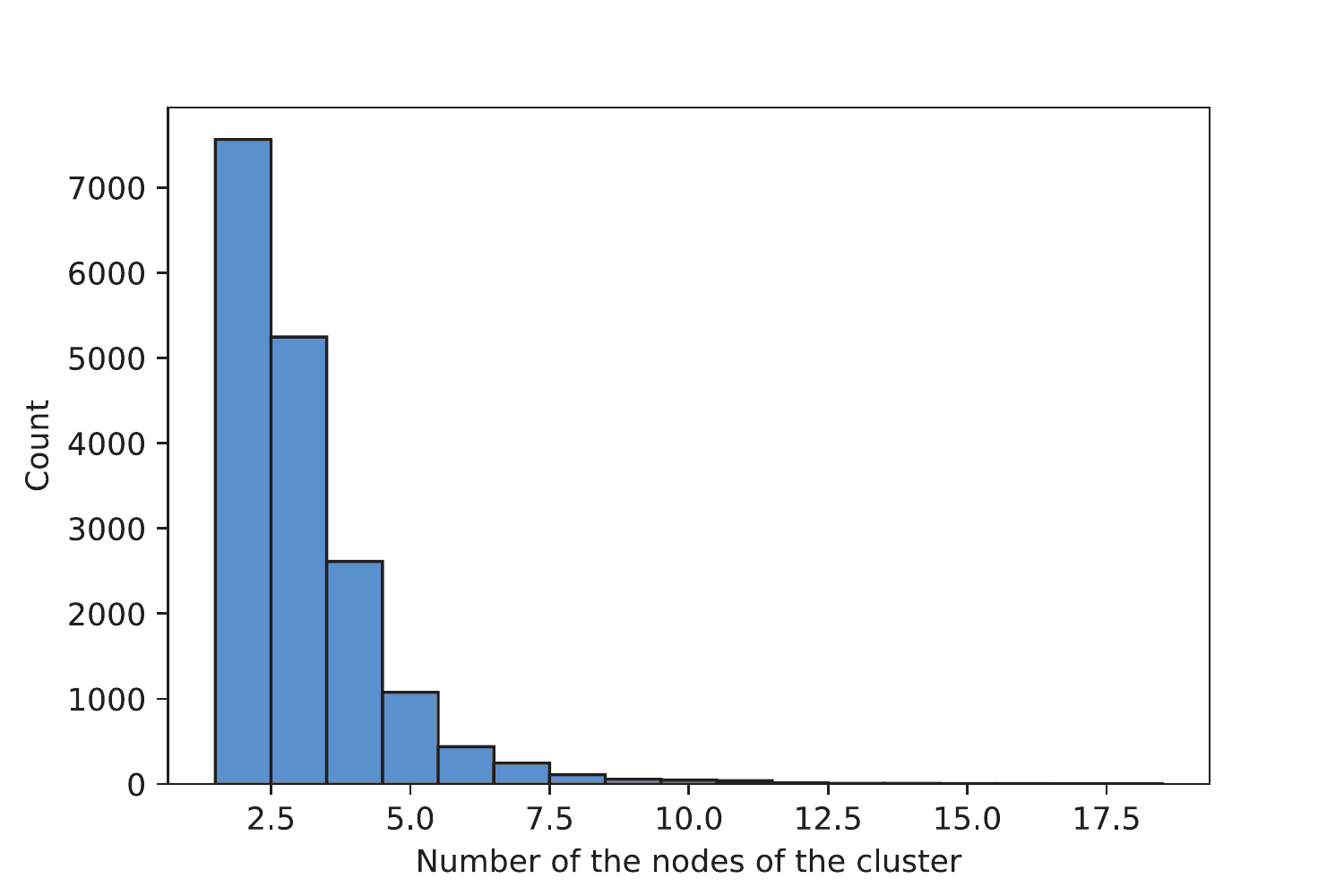}} 
\caption{The distribution of the number of the nodes in the subgraphs (\emph{i.e.} the size of the subgraphs). }
\label{stats:cluster-sizes}
\vspace{-0.1in}
\end{figure}

Figure~\ref{stats:cluster-sizes} shows the distribution of the number of the nodes in the subgraphs (\emph{i.e.} the size of the subgraphs). Note that the maximum size of the subgraphs in a graph is the so-called width of the corresponding tree decomposition of the graph. However, the average size of the clusters gives a better estimation of the number of the decisions required for generating the connections, compared to the width which gives an upper bound.

\subsection{Number of New Nodes in the Subgraphs}
\label{appendix:new-node-dist}
\begin{figure}[ht!]
\centering
\subfloat[]{\includegraphics[width = 1.4in]{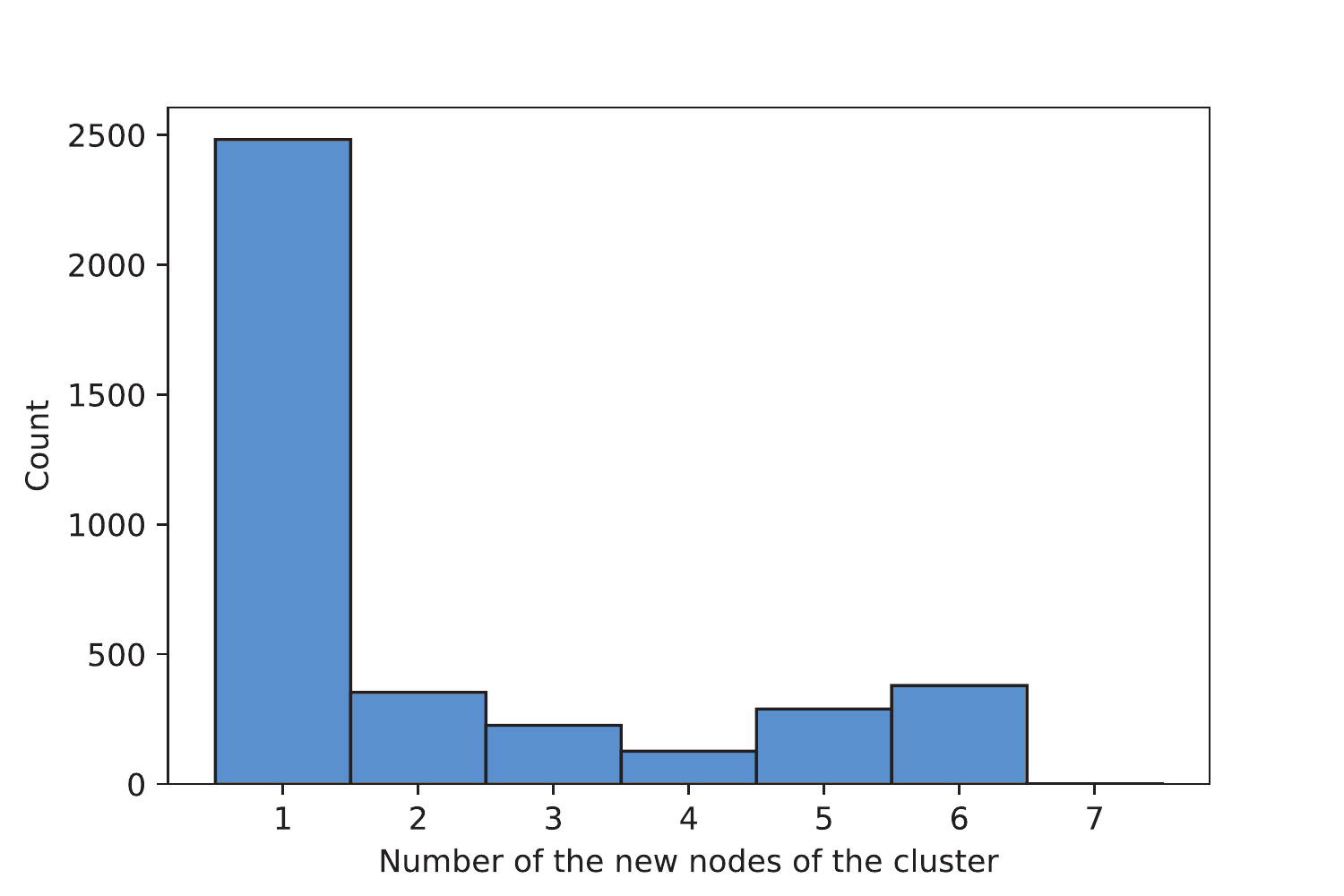}} 
\hspace{-0.1in}
\subfloat[]{\includegraphics[width = 1.4in]{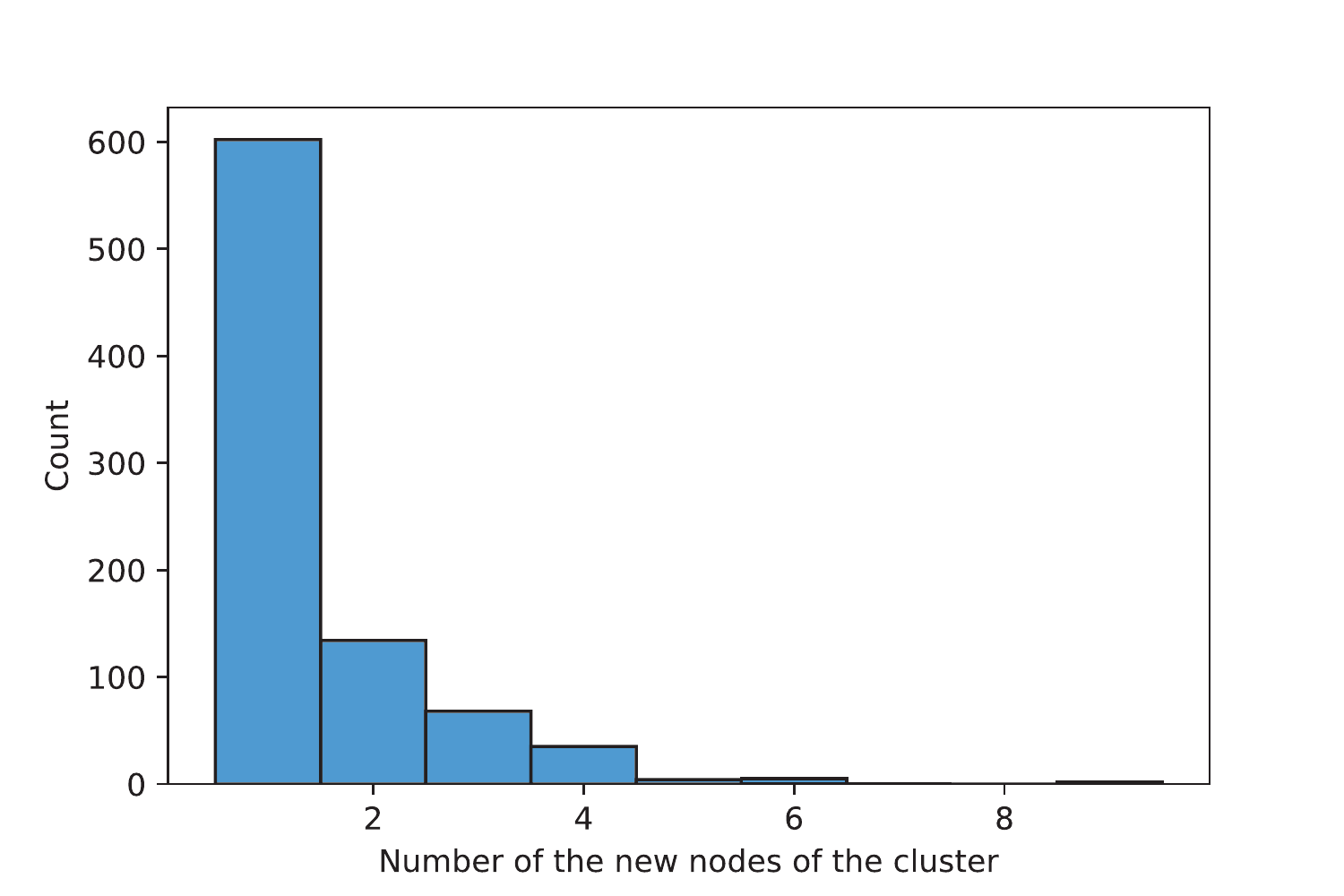}}
\hspace{-0.1in}
\subfloat[]{\includegraphics[width = 1.4in]{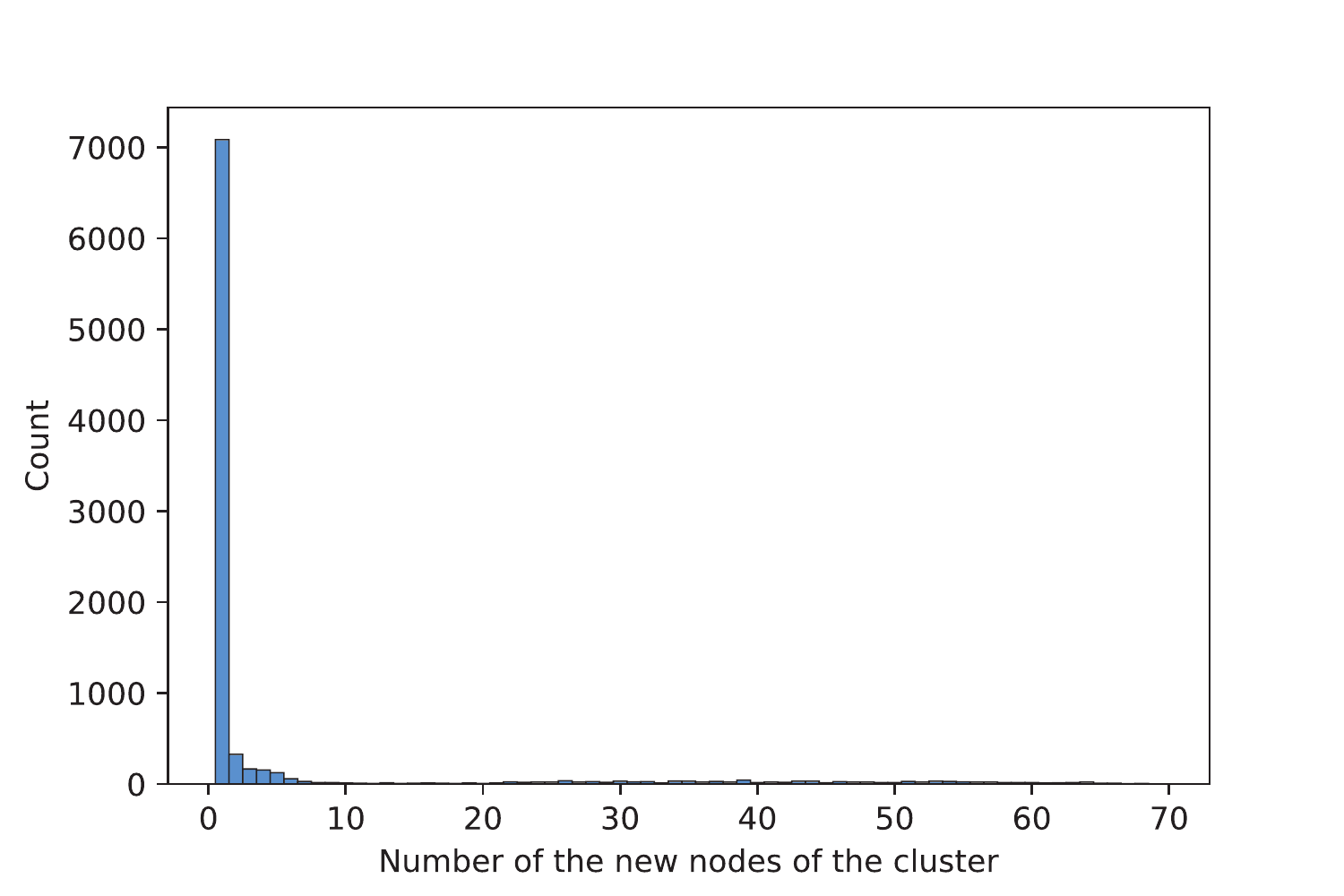}}
\hspace{-0.1in}
\subfloat[]{\includegraphics[width = 1.4in]{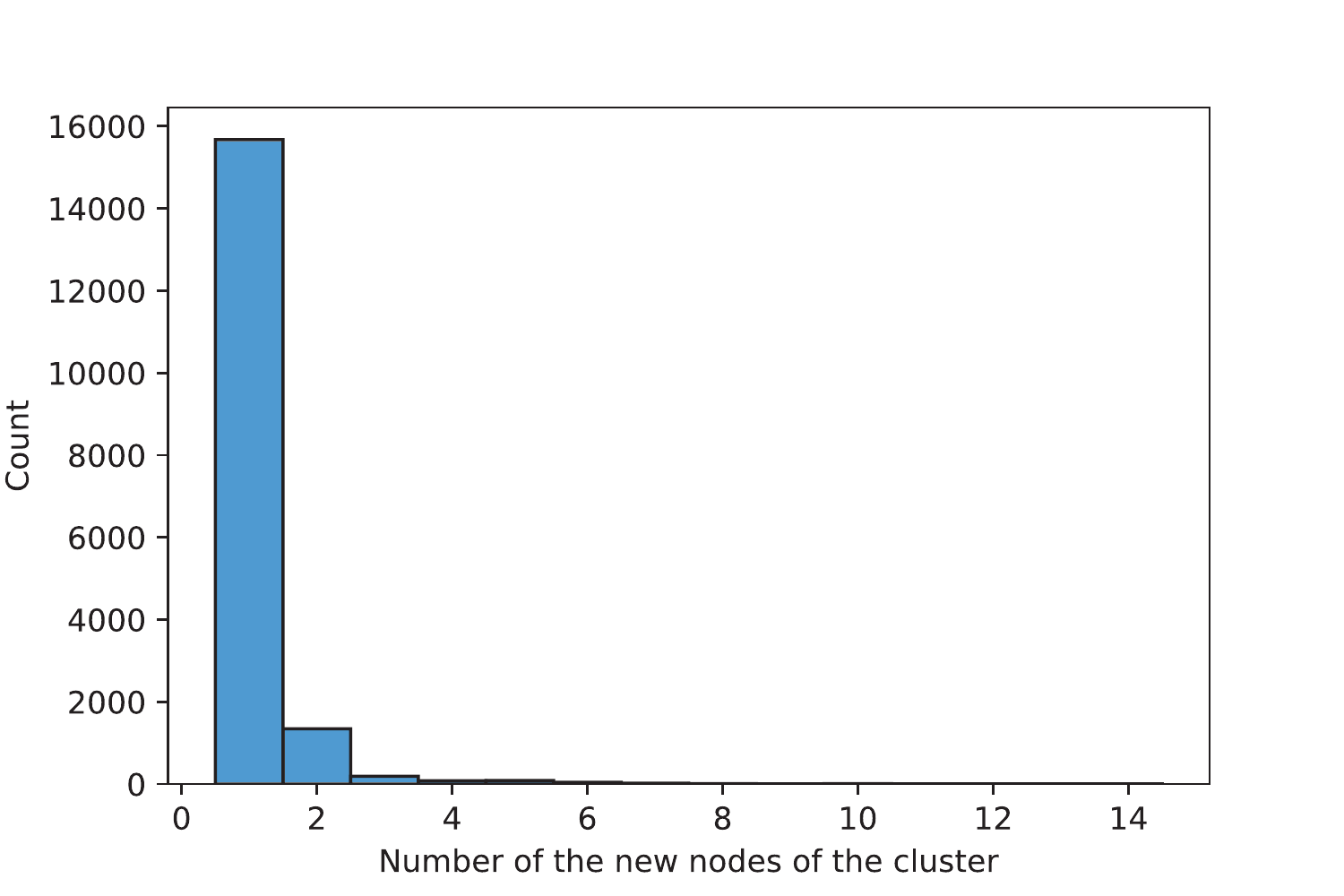}} 
\caption{The distribution of the number of the new nodes added in subgraph generation. }
\label{stats:new-nodes}
\vspace{-0.1in}
\end{figure}

Figure~\ref{stats:new-nodes} shows the distribution of the number of new nodes added in a subgraph. Interestingly, it can be observed that only one new node is added in most of the subgraphs. This verifies that the decision process for adding new nodes in a subgraph often has a low complexity.

\section{LOBSTER TREES EXPERIMENTS}
\label{sec:lobster}
Lobster trees are trees with a path backbone and all leaves have maximum distance two from the backbone. Previously, GRAN have conducted experiments on this dataset. Here we compare our tree generator model with JT-VAE and other graph generators on Lobster trees (Table~\ref{tab:lobster-res}). Dataset contains 100 synthetic lobster trees. Similar to Table~\ref{table:overfit}, two versions of the model with standard training (Tree-GEN) and overfitting (Tree-GEN*) are compared with state-of-the-art models. 
Fortunately, a measure of accuracy is also available for the Lobster dataset by counting the number of real Lobster trees out of 100 randomly generated samples. It can be observed that our proposed model can achieve the best results on all measures (except for degree) by overfitting on the training data. 
However, considering the likelihood-based evaluation, TD-GEN* verifies its overfitted behaviour with a substantially better performance on the train data, NLL$_{train} = 8.36$, compared to the test data, NLL$_{test} = 74.27$; 
however, the TD-GEN clearly does not suffer from such with NLL$_{train} = 28.80$ and NLL$_{test} = 28.79$.

\begin{table}[t]
\vspace{-0.3in}
\centering
\small
\setlength{\tabcolsep}{3pt}
\vskip 0.2in
\caption{
Tree generator models' performance on the Lobster tree dataset.
Tree-GEN is the proposed tree generator with early stopping with respect to 
the validation NLL. 
TD-GEN* is the same model overfitted to the training data. The final row shows the distance of the train dataset from the test dataset.
For statistical distances, the number of generated graphs is equal to the size of the test dataset.
}
\label{tab:lobster-res}
\scalebox{0.9}{
\begin{tabular}{@{}l|ccccc@{}}
\toprule
\multicolumn{1}{c|}{\multirow{2}{*}{Model}} & \multicolumn{5}{c}{Lobster Trees} \\ \cmidrule(l){2-6} 
\multicolumn{1}{c|}{} & Deg. & Clus. & Orbit. & Spec. & Acc. \\ \midrule
GraphVAE~\citep{simonovsky2018graphvae} & $2.09e^{-2}$ & $7.97e^{-2}$ & $1.43e^{-2}$ & $3.94e^{-2}$ & $0.09$ \\
GraphRNN-S~\citep{you2018graph} & $3.48e^{-3}$ & $4.30e^{-2}$ & $2.48e^{-4}$ & $6.72e^{-2}$ & $0.0$ \\
GraphRNN~\citep{you2018graph} & $\mathbf{9.26e^{-5}}$ & $\mathbf{0.0}$ & $2.19e^{-5}$ & $1.14e^{-2}$ & $\mathbf{1.0}$ \\
GRAN~\citep{liao2019efficient} & $3.73e^{-2}$ & $\mathbf{0.0}$ & $7.67e^{-4}$ & $2.71e^{-2}$ & $0.88$ \\
JT-VAE~\citep{jin2018junction}   & $0.163$      & $0.0$ & $5.86 e^{-3}$ & $8.27e^{-2}$ & $0.64$ \\ 
Tree-GEN [ours] & $2.94e^{-4}$ & $\mathbf{0.0}$ & $2.23 e^{-5}$ & $1.88e^{-2}$ & $0.99$ \\
Tree-GEN* [ours] & $2.07e^{-4}$ & $\mathbf{0.0}$ & $\mathbf{1.35e}^{\mathbf{-5}}$ & $\mathbf{5.44e}^\mathbf{-3}$ & $\mathbf{1.0}$ \\ \midrule
Train Dataset & $6.52e^{-4}$ & $0.0$ & $5.78e^{-5}$ & $0.024$ & $1.0$ \\ 
\bottomrule
\end{tabular}
}
\vspace{-0.2in}
\end{table}

\section{QUALITATIVE EXAMPLES}
\begin{figure*}[ht]
\centering
\subfloat[]{\includegraphics[width = 2.5in]{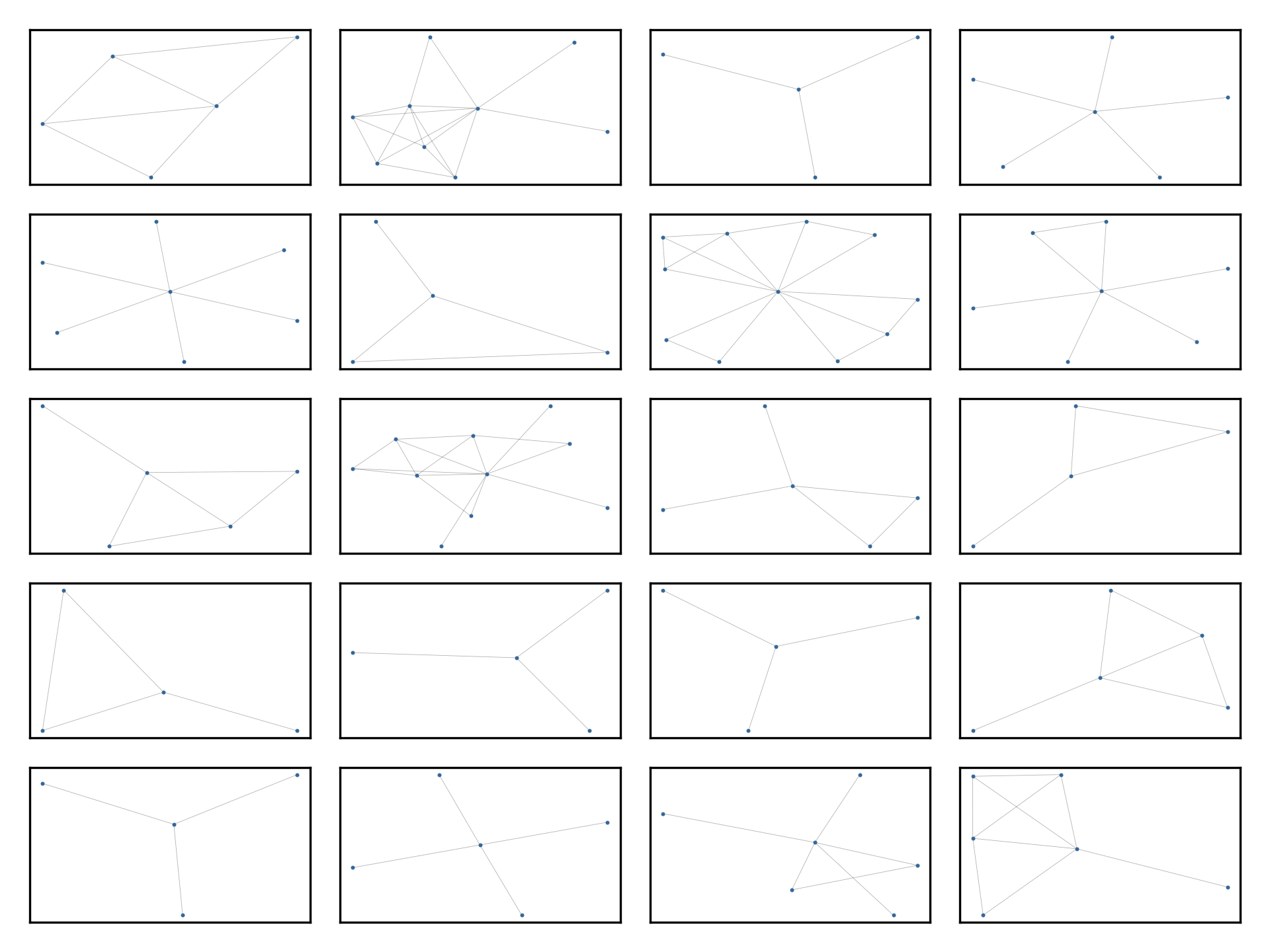}} 
\subfloat[]{\includegraphics[width = 2.5in]{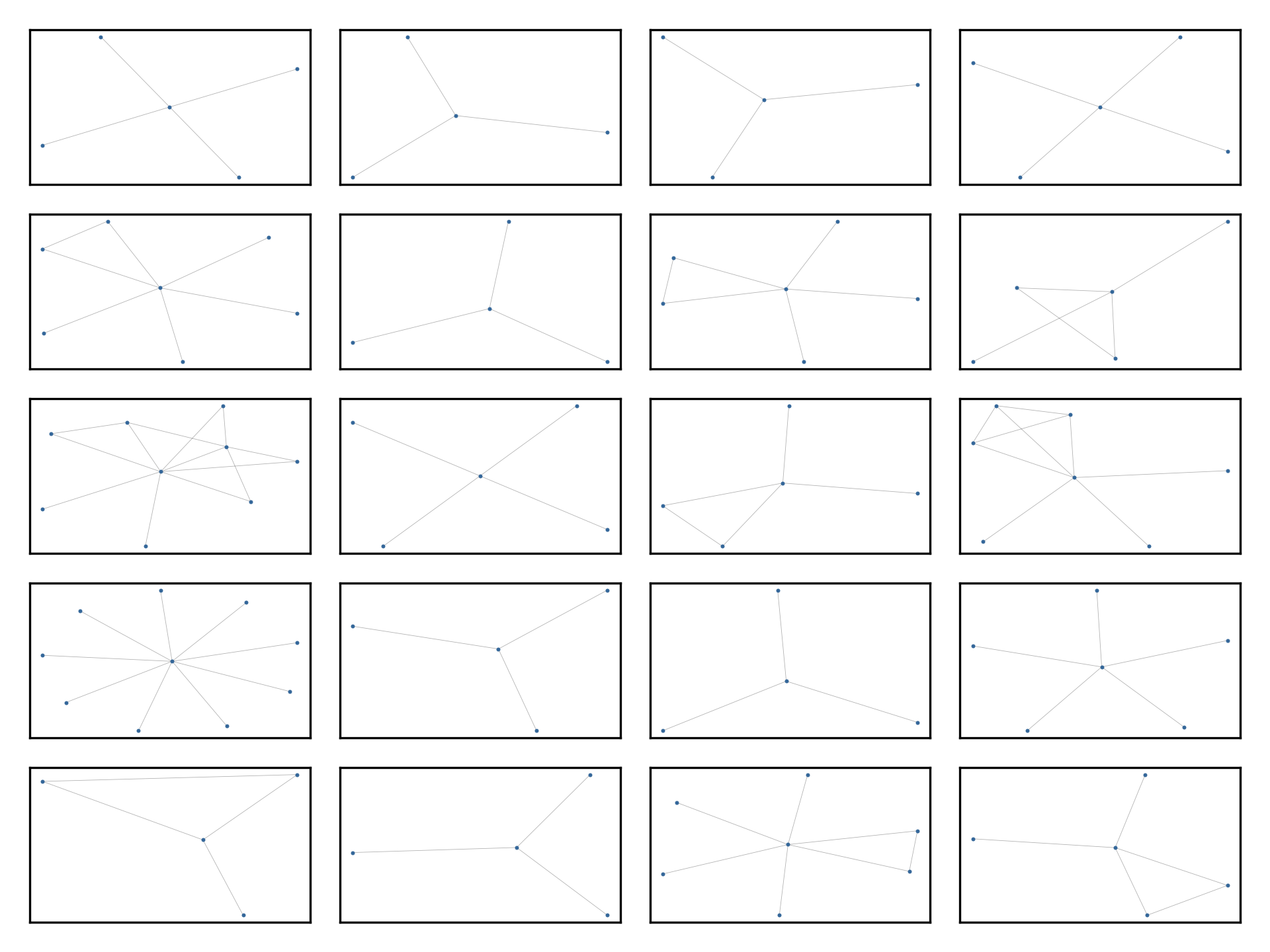}}
\caption{Qualitative examples for the Ego-small dataset. (a) Samples from the test dataset. (b) Samples generated from the standard training model.}
\label{fig:quality-ego}
\vspace{-0.3in}
\end{figure*}

\begin{figure*}[ht]
\centering
\subfloat[]{\includegraphics[width = 2.5in]{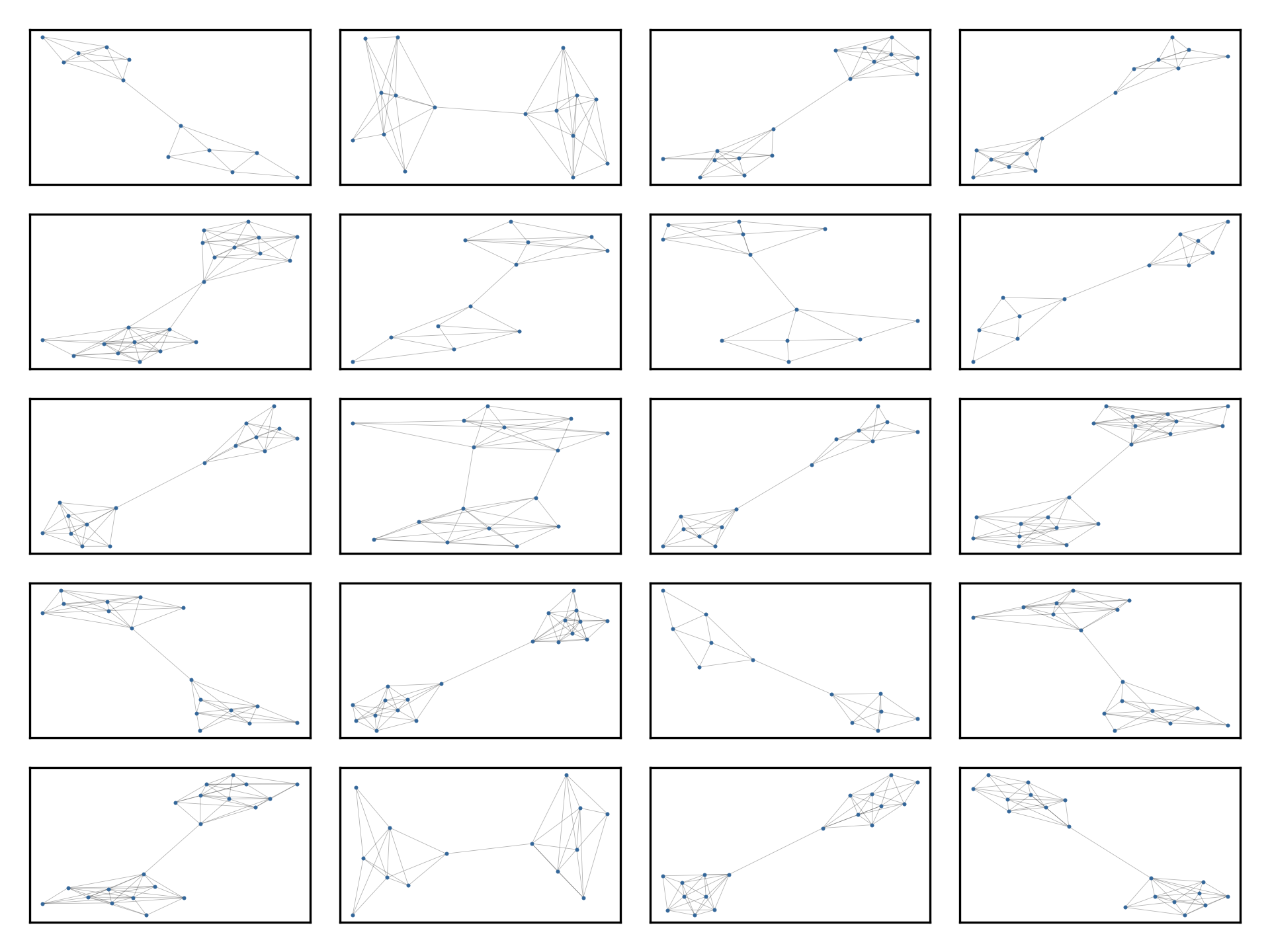}} 
\subfloat[]{\includegraphics[width = 2.5in]{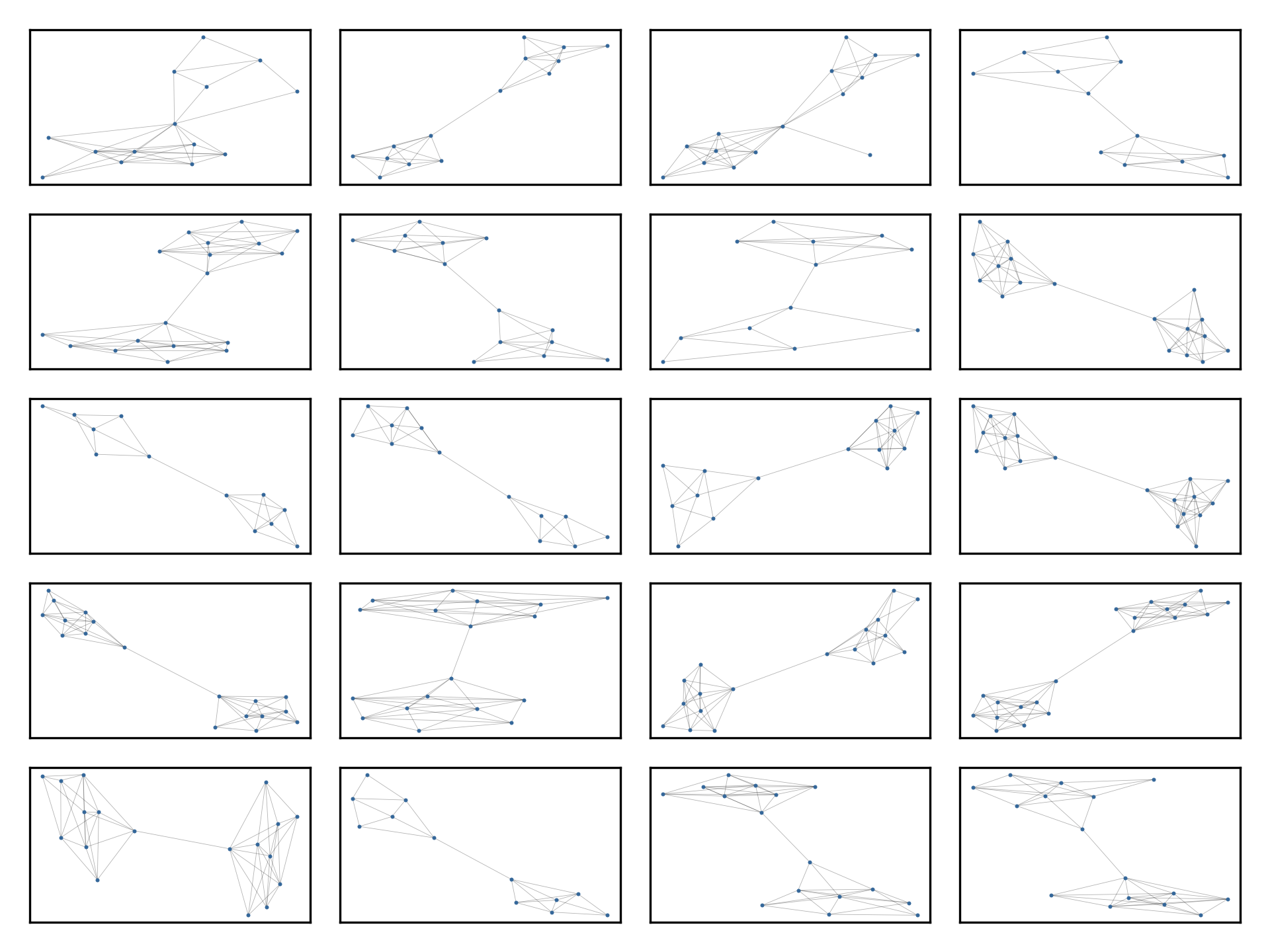}}
\caption{Qualitative examples for the Community-small dataset. (a) Samples from the test dataset. (b) Samples generated from the standard training model.}
\label{fig:quality-community}
\vspace{-0.3in}
\end{figure*}

\begin{figure*}[ht]
\centering
\subfloat[]{\includegraphics[width = 2.5in]{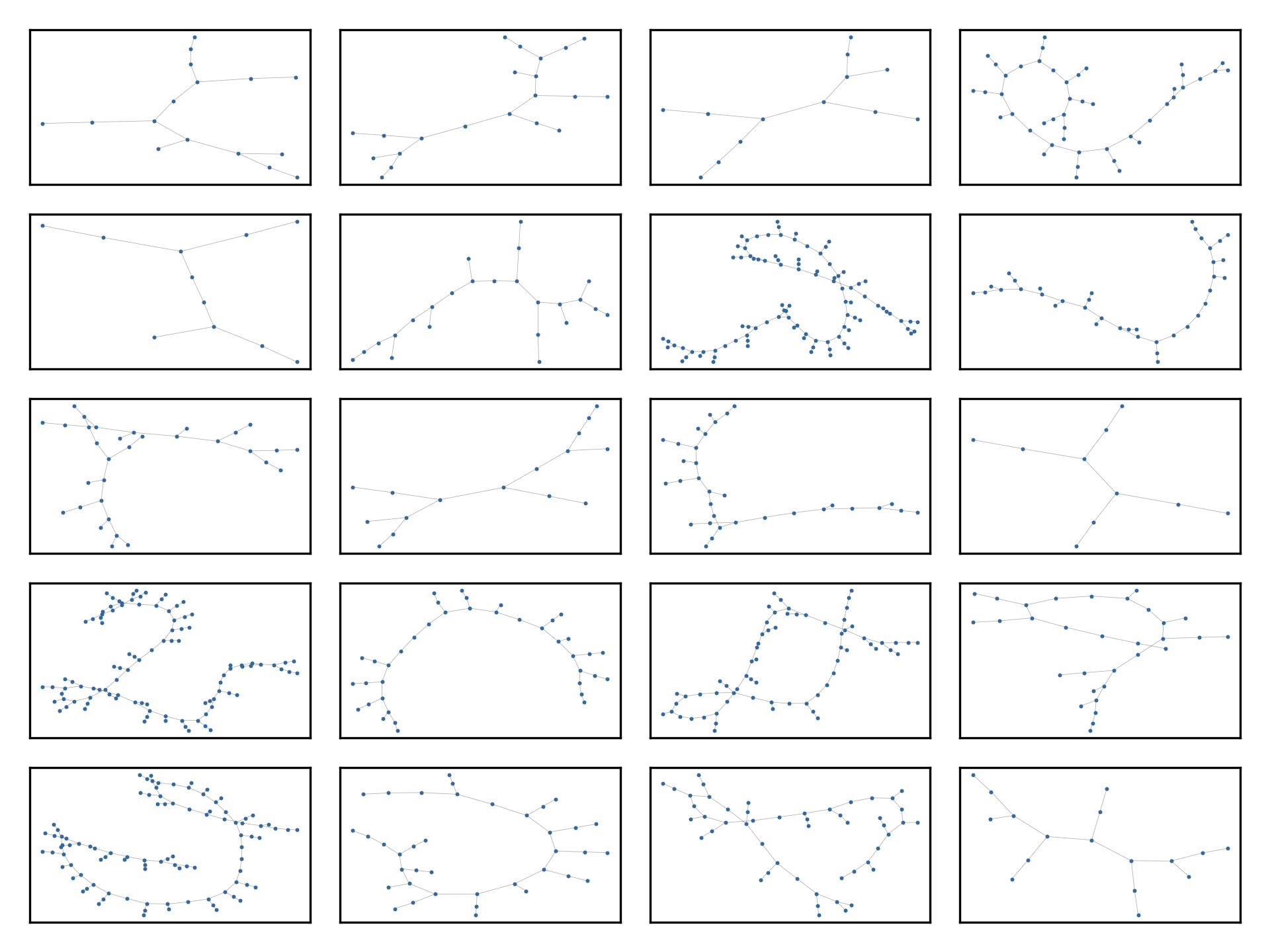}} 
\subfloat[]{\includegraphics[width = 2.5in]{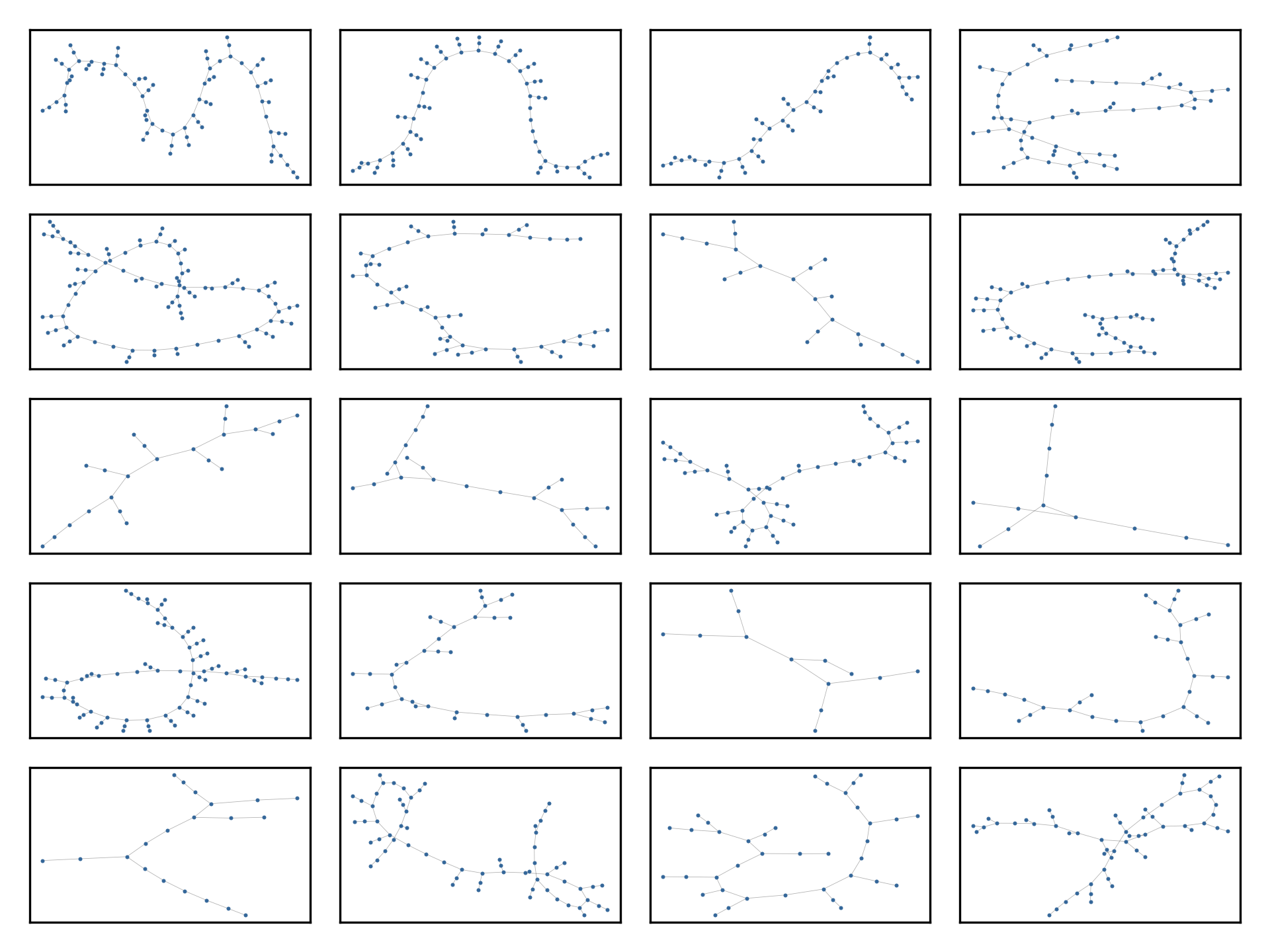}}
\caption{Qualitative examples for the Lobster trees dataset. (a) Samples from the test dataset. (b) Samples generated from the standard training model.}
\label{fig:quality-lobster}
\vspace{-0.3in}
\end{figure*}

In this section, we provide some qualitative examples generated by TD-GEN using the standard setting for training (\emph{i.e.} not overfitting). The examples are demonstrated in Figures~\ref{fig:quality-ego}, \ref{fig:quality-community}, and \ref{fig:quality-lobster} for the three datasets: Ego-small, Community-small, and Lobster trees. There are also some samples from the test graph datasets for comparison.

\end{document}